\documentclass{article} % For LaTeX2e
\usepackage[final]{colm2026_conference}

\usepackage{microtype}
\usepackage{graphicx}
\usepackage{subcaption}
\usepackage{booktabs} % for professional tables

\usepackage{hyperref}
\usepackage{url}

\usepackage{lineno}

\usepackage{amsmath}
\usepackage{amssymb}
\usepackage{mathtools}
\usepackage{amsthm}

\usepackage[capitalize,noabbrev]{cleveref}

\theoremstyle{plain}
\newtheorem{theorem}{Theorem}[section]

\newtheorem{corollary}[theorem]{Corollary}
\theoremstyle{definition}

\theoremstyle{remark}

\usepackage[textsize=tiny]{todonotes}

\usepackage{booktabs, multirow, makecell, array}
\usepackage{tikz, pgfplots, xcolor, amsmath}
\usepackage[T1]{fontenc} % For proper font encoding
\usepgfplotslibrary{groupplots}
\usepackage{pgfplotstable}
\pgfplotsset{compat=1.18}
\usepackage{wrapfig, caption}

\definecolor{teal1}{RGB}{77,175,159}
\definecolor{blue1}{RGB}{146, 181, 202}
\definecolor{red1}{RGB}{230, 145, 145}
\definecolor{orange1}{RGB}{255,158,74}
\definecolor{purple1}{RGB}{158,154,200}
\definecolor{green1}{RGB}{181,207,107}
\definecolor{deepblue}{RGB}{89, 156, 180}
\definecolor{deepred}{RGB}{194, 87, 89}
\definecolor{warmred}{RGB}{196,78,82}
\definecolor{lightgray}{RGB}{245,245,245}
\definecolor{darkgray}{RGB}{80,80,80}

\pgfplotsset{
    every axis legend/.append style={
        legend image post style={scale=0.3}
    }
}

\newcommand{\figfont}{\fontfamily{ppl}\selectfont} % Palatino

\pgfplotsset{every axis legend/.append style={font=\footnotesize\figfont}}

\numberwithin{equation}{section}

\usepackage{enumitem}
\usepackage{pifont}
\tikzset{
    stage/.style={
        rectangle, rounded corners=8pt,
        minimum width=3.5cm, minimum height=2.2cm,
        text width=3cm, text centered,
        draw=darkgray, line width=1.5pt,
        font=\footnotesize
    },
    input/.style={
        rectangle, rounded corners=6pt,
        minimum width=2.8cm, minimum height=1.8cm,
        text width=2.4cm, text centered,
        draw=teal1, fill=teal1!20, line width=2pt,
        font=\footnotesize
    },
    output/.style={
        rectangle, rounded corners=8pt,
        minimum width=3.8cm, minimum height=2cm,
        text width=3.4cm, text centered,
        draw=green1!80!darkgray, fill=green1!25, line width=2.5pt,
        font=\footnotesize
    },
    agent/.style={
        rectangle, rounded corners=4pt,
        minimum width=2.2cm, minimum height=1.4cm,
        text width=2cm, text centered,
        draw=deepblue, fill=deepblue!15, line width=1pt,
        font=\tiny
    },
    arrow/.style={
        ->, >=stealth, line width=2pt, darkgray
    },
    dashedarrow/.style={
        ->, >=stealth, line width=1.5pt, darkgray, dashed
    },
    title/.style={
        font=\Large\bfseries, darkgray
    },
    subtitle/.style={
        font=\normalsize\bfseries, warmred
    }
}

\title{Overconfident and Blind to Details: Fixing Prompt Insensitivity with Abductive Preference Learning}

\author{
Yijin Ni\textsuperscript{1} \quad
Simon Yu\textsuperscript{2} \quad
Peng Qi\textsuperscript{3} \\
\textsuperscript{1}Georgia Institute of Technology \quad
\textsuperscript{2}Northeastern University \quad
\textsuperscript{3}Uniphore \\
\texttt{\{yni64\}@gatech.edu}
}

\begin{document}

\ifcolmsubmission
\linenumbers
\fi

\maketitle

\begin{abstract}
Vision and language models frequently ignore semantically critical input edits, defaulting to pretraining priors. For example, models will confidently assert a five-legged dog has four legs; consequently, on the VLMBias benchmark, GPT 5.2 and Claude Sonnet 4.6 achieve only $4.6\%$ and $0\%$ accuracy, respectively. Existing methods address this problem through building up datasets that covers the underrepresented inputs to tune the policy function $\pi(y \mid x)$, where $x$ and $y$ refer to input prompts and responses, respectively. 
However, prompting baselines yield gains of under $3\%$ on VLMBias due to the low probability density of rare prompts. To bypass this bottleneck, we propose \emph{abductive preference learning} to optimize the abductive policy $\pi(x \mid y)$. We prove this amplifies forward policy improvements by a factor of $q(y)/p(x)$, where $p(\cdot)$ and $q(\cdot)$ denote the marginal probabilities of the prompt and response, yielding the largest gains on the rarest prompts. 
Furthermore, we demonstrate that for translation invariant pairwise preference learning methods, such as DPO, estimating $\pi(x \mid y)$ reduces to a structural data swap that compares prompts for a fixed response, requiring no architectural changes. 
Empirically, abductive preference learning delivers large gains on long-tail prompt sensitivity: on VLMBias, A-DPO raises accuracy from $3\%$ to $44\%$ ($14\times$), outperforming GPT-5.2 ($4.6\%$) and all closed-source VLMs except Gemini~3~Flash; on Inverse-IFEval, Multi-DPOP reaches $65$--$84\%$, surpassing GPT-5 ($73.7\%$) at the 9B scale while preserving IFBench, unlike DPO which degrades it by $8$--$12\%$.

\end{abstract}

\section{Introduction}
\label{sec:intro}

\begin{figure}[ht]
\centering
\begin{minipage}[c]{0.48\linewidth}
\centering
\resizebox{\linewidth}{!}{
\small
\begin{tabular}{p{0.4\linewidth} p{0.6\linewidth} p{0.15\linewidth}}
\toprule
\textbf{General prompt} & \textbf{Specified prompt} & \textbf{GPT-4} \\
\midrule
Where was the \textcolor{black}{last Olympics} held? & \textcolor{black}{I am writing a novel describing life on }\textcolor{deepred}{\bf Mars} \textcolor{black}{.}
Where was the last Olympics held \textcolor{deepred}{\bf in my text}? & Paris \\
\midrule
Can I eat the \textcolor{deepblue}{\bf food} that has been left out overnight? & Can I eat the \textcolor{deepred}{\bf potato chips} that have been left out overnight? & No \\
\bottomrule
\end{tabular}
}
\end{minipage}%
\hfill
\begin{minipage}[c]{0.5\linewidth}
\centering
\includegraphics[width=\linewidth]{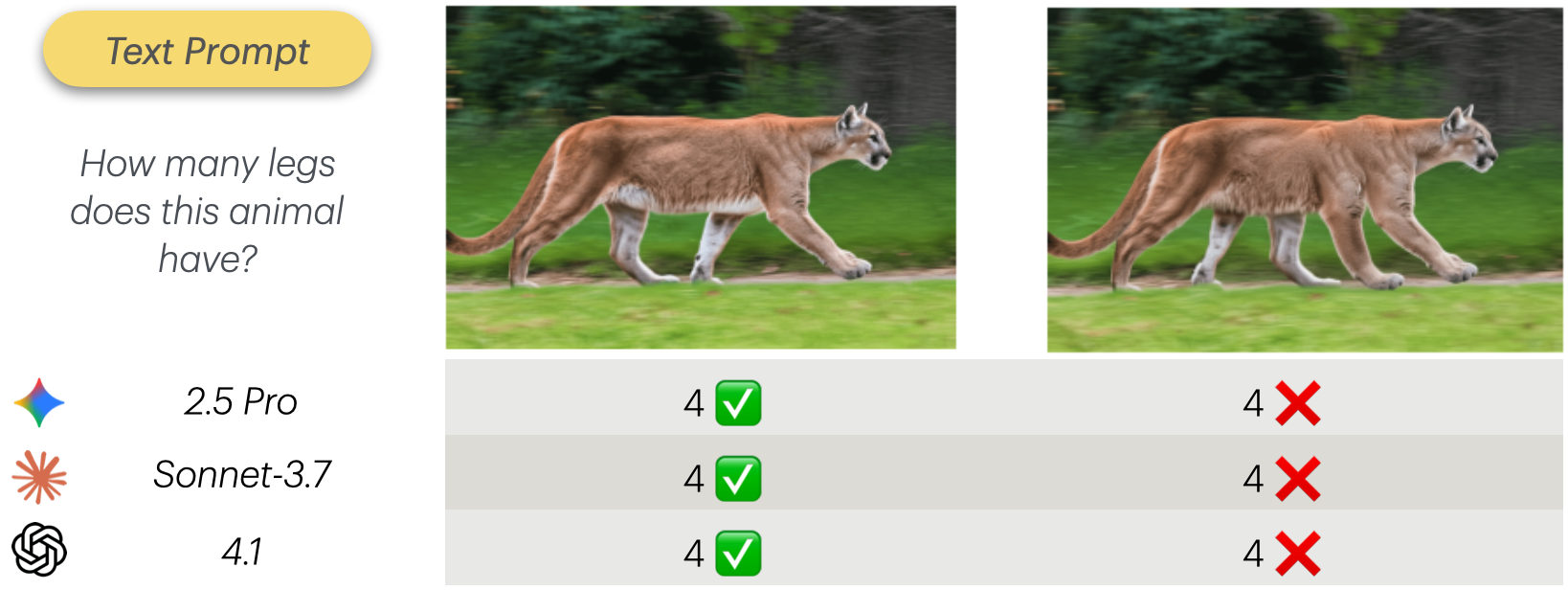}
\end{minipage}
\caption{Illustrative examples of the long-tail prompt insensitivity phenomenon.}
\label{fig:overconfidence_combined}
\end{figure}

%% ---- Para 1: Problem formulation — counterfactual prompt sensitivity ----
Modern foundation models have been shown to be insensitive to semantically crucial changes in the prompt, leading to undesirable responses.
For instance, vision-language models can confidently report that a dog has four legs even when the image clearly shows five~\citep{vo2025visionlanguagemodelsbiased}. Subtle edits to an image that flip whether a scene is humorous are invisible to all tested models, which perform at chance~\citep{jain2024ai}. These failures persist across frontier systems: on the VLMBias benchmark, GPT-5.2 achieves $4.6\%$, and Claude Sonnet 4.6 achieves {${0\%}$} on counterfactual visual perception (Table~\ref{tab:vlm_comprehensive}).
In the text domain, base models score only $16$--$31\%$ on the counterintuitive instruction-following constraints of \textsc{Inverse-IFEval}~\citep{zhang2025inverseifevalllmsunlearn}.
Figure~\ref{fig:overconfidence_combined} shows representative examples.

We formalize this as a \emph{long-tail input sensitivity} problem: given a general input $x_g$ and a specific input $x_s$ differing by a small but semantically critical edit, models produce $\pi(y_g \mid x_g) \approx \pi(y_g \mid x_s) \gg \pi(y_s \mid x_s)$, defaulting to the response associated with the more common input.
Prior work traces this to data imbalance: model accuracy correlates with training-corpus frequency~\citep{kandpal2023large}.
Existing approaches that augment data for $\pi(y \mid x)$ yield only marginal improvements~\citep{vo2025visionlanguagemodelsbiased}.

We introduce a training framework targeting abductive policy $\pi(x \mid y)$, the conditional probability of prompt $x$ given response vector $y$, which leverages the low empirical density of $x_s$ in the pretraining distribution as a learning advantage.
In contrast to the standard optimization of $\pi(y \mid x)$, where the long-tail prompt $x_s$ serves as the conditioning input, our approach conditions on the target response $y_s$.
When the response $y_s$ is not itself rare -- e.g., common tokens such as ``yes,'' ``no,'' a short count, or a factual answer -- the response has a high marginal probability $q(y_s)$.
Combined with a small $p(x_s)$, 
% Notably, $\pi(y \mid x)$ conditions on the rare $x_s$ with small prior $p(x_s)$, and forward training therefore faces the low-density bottleneck identified above.
% In contrast, $\pi(x \mid y)$ conditions on the common $y_s$ with large $q(y_s)$, bypassing this bottleneck entirely.
our theoretical result (Theorem~\ref{thm:amplification}) shows that improving $\pi(x_s \mid y_s)$ by $\delta$ increases the forward policy $\pi(y_s \mid x_s)$ by $\delta \cdot q(y_s)/p(x_s) \gg \delta$. %When $p(x_s) \ll q(y_s)$, this gain exceeds $\delta$.
Since the amplification factor $q(y_s)/p(x_s)$ increases as $p(x_s)$ decreases, optimizing $\pi(x \mid y)$ will yield the largest improvements on the rarest prompts.
Moreover, we prove that optimizing $\pi(x \mid y)$ reduces to swapping data roles in any translation-invariant preference method (DPO, IPO, GPO, SLiC-HF; Theorem~\ref{prop:general}), extend this to contrastive losses (Corollary \ref{cor:infonce}) and DPOP with bounded error (Corollary~\ref{cor:dpop_bound}).
Abductive preference learning outperforms all forward baselines on long-tail prompt sensitivity across both modalities while preserving general capabilities.
On \textsc{HumorDB} (visual humor detection), A-DPO raises pairwise accuracy from chance (${\sim}50\%$) to $85\%$, while GRPO, trained on the same data with a task reward, reaches only $61\%$ (Table~\ref{tab:vlm_comprehensive}).
On \textsc{VLMBias} (counterfactual visual perception), A-DPO improves from $3\%$ to $44\%$ (Qwen3-VL-8B), a $14\times$ gain, while GRPO reaches only $12\%$.
These results hold across five VLM families, and general capabilities (\textsc{MMMU-Pro}, \textsc{HallusionBench}) are preserved or improved (Appendix Table~\ref{tab:vlm_general}).
In the text domain, Multi-DPOP achieves $99.5\%$ standard accuracy and $85.0\%$ abductive accuracy on \textsc{A-HaluEval}, compared to the base model's $90.0\%$ / $54.7\%$ and GRPO's $97.5\%$ / $58.3\%$ (Table~\ref{tab:additional_summary}).
On \textsc{Inverse-IFEval}, Multi-DPOP reaches $\mathbf{65}$--$\mathbf{84\%}$ accuracy (from $33$--$49\%$ base), surpassing GPT-5 ($73.7\%$) at the 9B scale, while preserving general instruction following on IFBench.
Gains are consistent across five LLM families (Tulu-3.1, Gemma-3, GLM-4, Qwen3-4B, Qwen3-30B) and four model scales (4B--30B), with general benchmarks (\textsc{AlpacaEval-2}, \textsc{MMLU-Pro}, \textsc{GPQA}) preserved within $\pm 1\%$.

%% ---- Para 6: Contributions (3 items) ----
In summary, our contributions are:
\begin{enumerate}[leftmargin=*]
    \item \textbf{Abductive preference learning.}
    We propose estimating $\pi(x \mid y)$ instead of $\pi(y \mid x)$ for long-tail prompt sensitivity, and prove that improvements in $\pi(x \mid y)$ are amplified by $q(y)/p(x)$ in the forward direction (Theorem~\ref{thm:amplification}).

    \item \textbf{Role-switch equivalence.}
    We prove that estimating $\pi(x \mid y)$ reduces to swapping data roles for any translation-invariant pairwise method, with no new architecture (Theorem~\ref{prop:general}), extending it to contrastive losses (Corollary \ref{cor:infonce}) and DPOP with bounded error (Corollary~\ref{cor:dpop_bound}).

    \item \textbf{Experimental validation.}
    On four benchmarks across text and vision, abductive methods substantially outperform all forward baselines and closed-source models: $50\% \to 85\%$ on \textsc{HumorDB}, $3\% \to 44\%$ on \textsc{VLMBias}, $55\% \to 92\%$ on \textsc{A-HaluEval}, and $49\% \to 84\%$ on \textsc{Inverse-IFEval} (surpassing GPT-5), with general capabilities preserved across five LLM and five VLM families.
\end{enumerate}

\section{Related Works}

\paragraph{Long-tail input sensitivity is documented across modalities and model scales.}
VLMs default to memorized priors rather than grounding in image content~\citep{vo2025visionlanguagemodelsbiased}, and subtle image edits that flip semantic meaning are invisible to all tested models~\citep{jain2024ai}.
In the text domain, LLM accuracy on factual questions degrades sharply for entities rare in the training corpus~\citep{kandpal2023large}, instruction-following benchmarks reveal stubborn adherence to memorized patterns~\citep{zhang2025inverseifevalllmsunlearn, mckenzie2023inverse, pyatkin2025generalizing}, and the reversal curse~\citep{berglund2023reversal} demonstrates a fundamental directional asymmetry in learned associations.
These failures are theoretically grounded in spectral bias~\citep{rahaman2019spectral, ronen2019convergence}, gradient starvation~\citep{pezeshki2021gradient}, and low-rank simplicity bias~\citep{huh2021low, saxe2013exact}, which cause neural networks to prioritize common patterns over rare details.
Efforts to mitigate these failures face a learning bottleneck caused by the rarity of $x_s$ in pretraining data~\citep{kandpal2023large}: the model assigns long-tail inputs such low probability density that correcting $\pi(y \mid x_s)$ requires a disproportionately large volume of training examples to override the initial prior.
On the VLMBias benchmark, instructing models to verify answers or rely strictly on image details yields improvements of only $+2.70$ and $+1.87$ percentage points over a $17\%$ baseline~\citep{vo2025visionlanguagemodelsbiased}.

\paragraph{Existing approaches augment data for $\pi(y \mid x)$ but remain within the forward paradigm.}
CF-VLM~\citep{zhang2025cfvlm} generates counterfactual image-text pairs by editing visual attributes (e.g., object color or count) and fine-tunes the model via SFT on both the factual and counterfactual pairs.
S-VCO~\citep{wu2025svco} decomposes the multimodal prompt into image and text components and trains the forward policy via DPO-style sub-losses that compare image-present versus image-absent conditions.
Both methods operate within $\pi(y \mid x)$ and are specific to VLMs.

\paragraph{Existing preference learning frameworks tuning forward policy $\pi(y \mid x)$.}
$\beta$-DPO~\citep{wu2024beta} adjusts temperature based on reward margins;
online DPO~\citep{qi2024online} generates on-policy data to address distribution shift;
DPOP~\citep{pal2024smaug} prevents preferred-response likelihood collapse;
and \citet{liu2024provably} show that SFT regularization mitigates reward overoptimization.
All of these methods target data quality or training dynamics within the forward paradigm $\pi(y \mid x)$ and do not address the conditioning direction.

% \paragraph{The directional asymmetry of autoregressive learning is documented but unexploited.}
% The reversal curse~\citep{berglund2023reversal} shows models trained on ``A is B'' fail to learn ``B is A,'' and \citet{zhu2024reversal} prove this is inherent to forward training.
% We exploit this asymmetry rather than diagnosing it.

\section{Abductive Preference Learning}
\label{sec:motivation}

In this section, we first prove that highly skewed pretraining distributions induce gradient starvation for fine-grained features, necessitating post-training intervention, and that forward preference learning ignores the detail feature even under long-tail data augmentation (Section~\ref{sec::linear_model}).
We then demonstrate that optimizing the abductive policy $\pi(x \mid y)$ leads to a provable gradient amplification factor $q(y_s)/p(x_s) \gg 1$ and guarantees recovery of the starved features, regardless of the dataset employed in preference learning (Section~\ref{sec:why_inverse}).
Finally, we establish that for any preference algorithm with a translation-invariant outer loss, the $\pi(x \mid y)$ objective reduces to an exact transposition of the contrastive inputs, with no architectural modification (Section~\ref{sec:role_switch}).

\subsection{Detail Starvation in Forward Training}
\label{sec::linear_model}

Throughout, $p(x)$ denotes the population prompt marginal; for any policy $\pi$, $q_\pi(y) := \sum_{x'} \pi(y \mid x')\, p(x')$ is the response marginal it induces, with $q_\theta$ and $q_{\mathrm{ref}}$ the instances for the trained policy $\pi_\theta$ and the frozen reference policy $\pi_{\mathrm{ref}}$ of preference learning. We abbreviate $q := q_\pi$ when the policy is clear from context; Appendix~\ref{app:notation} provides a summary table.

\begin{theorem}[Detail Suppression in Pre-Training]
\label{thm:starvation}
Let $\pi_\theta(y \mid x)$ be an autoregressive model producing response $y = (y_1, \ldots, y_T)$ given prompt embedding $x \in \mathbb{R}^d$, with each per-position log-probability $\log \pi_\theta(y_t \mid x, y_{<t})$ twice-differentiable in $(\theta, x)$.
Let $\mathcal{X}$ and $\mathcal{Y}$ denote the full prompt and response spaces.
Consider two subpopulations: frequent general prompts $x_g = v_c$ with large marginal $p(x_g)$ and response $y_g$, and rare specific prompts $x_s = v_c + v_d$ with $p(x_s) \ll p(x_g)$ and response $y_s = y_g + \delta_y$, where $v_c$ is the common feature shared by both subpopulations, $v_d$ is the detail feature with $\|v_d\|$ small, and $\delta_y$ is the response shift it induces.
Let the training dataset be partitioned into a general subset $\mathcal{D}_g$ of $N_g$ examples and a specific subset $\mathcal{D}_s$ of $N_s$ examples, with $N_g \gg N_s$ reflecting $p(x_g) \gg p(x_s)$.

The gradient of the pre-training loss $\mathcal{L} = -\sum_i \log \pi_\theta(y_i \mid x_i)$ decomposes as:
\begin{equation}
\label{eq:mle_grad}
    \nabla_\theta \mathcal{L} = \underbrace{\sum_{i \in \mathcal{D}_g \cup \mathcal{D}_s} \nabla_\theta \log \pi_\theta(y_i \mid v_c)}_{\text{common: }O(N_g + N_s)} \;+\; \underbrace{\sum_{i \in \mathcal{D}_s} \bar{\mathbf{H}}_\theta(v_c, y_i)\, v_d}_{\text{detail: }O(N_s)} \;+\; O(\|v_d\|^2),
\end{equation}
where $\bar{\mathbf{H}}_\theta(x, y) := \sum_{t=1}^{T} \frac{\partial^2 \log \pi_\theta(y_t \mid x, y_{<t})}{\partial \theta\, \partial x^T}$ is the aggregate mixed Hessian over all token positions.
Under frequency imbalance $N_g \gg N_s$, the detail component is a vanishing fraction of the total gradient.
\end{theorem}
All proofs are deferred to Appendix~\ref{app:proofs}.
For VLMs, $v_d = x_w - x_l$ is exact by construction (e.g., two images differing by a subtle edit), where $x_w$ and $x_l$ denote the preferred and rejected prompts of an abductive triple $(x_w, x_l, y)$.

Theorem~\ref{thm:starvation} applies to pre-training, where the data distribution is fixed and inherently imbalanced.
Rebalancing web-scale pre-training corpora is not feasible, so post-training must correct the resulting detail-feature deficit.
Existing post-training methods address this by constructing balanced fine-tuning data and training $\pi(y \mid x)$ on it.
However, even GRPO with balanced on-policy data and task-specific rewards achieves substantially lower long-tail prompt sensitivity than our method across all benchmarks (Section~\ref{sec:expanded_vlm}).

The following result shows that this gap is structural, not a matter of data quality.
Even in the best case for the forward paradigm, where every training sample is an augmented specific prompt, the per-sample gradient remains dominated by a detail-blind component.

\begin{theorem}[Detail-Blindness of Forward Preference Learning]
\label{thm:forward_balanced}
Let $\psi(x, y) := \log \pi_\theta(y \mid x) - \log \pi_{\mathrm{ref}}(y \mid x)$ be the log-likelihood ratio against a frozen reference model, and let $F$ be any translation-invariant outer function, i.e., $F(a+c, b+c) = F(a,b)$ for all $c$; this class includes DPO, IPO, SLiC-HF, and GPO.
Consider the forward pairwise loss $\mathcal{L}_{\mathrm{fwd}} = -\sum_{i=1}^{N} F(\psi(x_i, y_{w,i}),\, \psi(x_i, y_{l,i}))$, where $y_{w,i}$ and $y_{l,i}$ denote the preferred and rejected responses for prompt $x_i$, trained on \textbf{long-tail-augmented} data: all $N$ prompts are specific, i.e., $x_i = v_c + v_d$ with $N_s = N$ and $N_g = 0$.
The gradient decomposes as:
\begin{equation}
\label{eq:forward_balanced}
\begin{aligned}
    \nabla_\theta \mathcal{L}_{\mathrm{fwd}} = -\sum_{i=1}^{N} F'(s_i) \cdot \underbrace{\Big[\nabla_\theta \log \pi_\theta(y_{w,i} \mid v_c) - \nabla_\theta \log \pi_\theta(y_{l,i} \mid v_c)\Big]}_{\text{response discrimination at } v_c\text{: } O(1),\; v_d\text{-blind}}\\
    \;-\; \sum_{i=1}^{N} F'(s_i) \cdot \underbrace{\Big[\bar{\mathbf{H}}_\theta(v_c, y_{w,i}) - \bar{\mathbf{H}}_\theta(v_c, y_{l,i})\Big] v_d}_{\text{detail-sensitive: } O(\|v_d\|)} \;+\; O(\|v_d\|^2),
\end{aligned}
\end{equation}
where $F'(s_i) := \partial F / \partial a \big|_{(a,b)=(\psi_w, \psi_l)}$ is a scalar weight, with $\psi_w := \psi(x_i, y_{w,i})$ and $\psi_l := \psi(x_i, y_{l,i})$.
Because $F'$ multiplies both components of the same sample equally, the per-sample detail fraction is $O(\|v_d\|) \ll 1$, independent of the data composition $N_s / N_g$.
\end{theorem} Proof in Appendix~\ref{app:forward_balanced}.

\textbf{Comparison with abductive training.}
Under the backward loss (Theorem~\ref{thm:recovery}), each sample compares two prompts $x_w$ and $x_l$ given the same response $y$.
The zeroth-order terms $\nabla_\theta \log \pi_\theta(y \mid v_c)$ cancel exactly, leaving only the $O(\|v_d\|)$ detail component.
Thus backward training directs its gradient toward $v_d$, while forward training directs only an $O(\|v_d\|)$ fraction, even with perfectly balanced data.
This is the structural reason why data augmentation within the forward paradigm cannot match abductive training.

\subsection{From $\pi(y \mid x)$ to $\pi(x \mid y)$: Overcoming Data Sparsity for Detail Features}
\label{sec:why_inverse}

We now show that learning $\pi(x \mid y)$ instead of $\pi(y \mid x)$ turns the imbalance $p(x_s) \ll p(x_g)$ of Theorem~\ref{thm:starvation} from a weakness into an advantage.

\begin{theorem}[Inverse Amplification]
\label{thm:amplification}
Let $\pi(y \mid x)$ and $\pi(x \mid y)$ be related by Bayes' rule:
\begin{equation}
\label{eq:bayes_amplify}
    \pi(y \mid x) = \frac{\pi(x \mid y)\, q(y)}{p(x)}.
\end{equation}
Consider a specific prompt $x_s$ with target response $y_s$, and suppose $p(x_s) \ll q(y_s)$.
If learning $\pi(x \mid y)$ improves $\pi(x_s \mid y_s)$ by $\delta > 0$, i.e., $\pi'(x_s \mid y_s) = \pi(x_s \mid y_s) + \delta$, then the induced change in the forward conditional satisfies:
\begin{equation}
\label{eq:amplification}
    \Delta_{\pi(y_s \mid x_s)} = \delta \cdot \frac{q(y_s)}{p(x_s)}\!\left(1 + O\!\left(\frac{p(x_s)}{q(y_s)}\right)\right).
\end{equation}
In particular, the amplification factor $A := q(y_s)/p(x_s) \gg 1$.
\end{theorem}
The proof is in Appendix~\ref{app:proof_amplification}.

\textit{Remark (When does $p(x_s) \ll q(y_s)$ hold?).}
The condition is naturally satisfied when specific prompts are rare but their responses are generic: in \textsc{HumorDB} the response is ``Yes'' or ``No''; in \textsc{VLMBias}, ``4'' or ``5''; in \textsc{A-HaluEval}, a short factual answer.
Many different prompts produce the same response $y_s$, so $q(y_s)$ is large while $p(x_s)$ is small.
In the visual domain, $p(x_s)$ can be exponentially small in the effective dimensionality of the image representation, making the amplification factor $A$ particularly large.

\textbf{The imbalance that causes starvation under $\pi(y \mid x)$ enables amplification under $\pi(x \mid y)$.}
Forward training fails because the gradient for $v_d$ scales with $p(x_s)$, which is small.
Under $\pi(x \mid y)$, the roles of prompts and responses are reversed: the ``input'' to the inverse model is the response $y_s$, which is generic and therefore frequent ($q(y_s)$ is large).
The same data imbalance that starves forward training ($p(x_s)$ small) amplifies inverse training (the ratio $q(y_s)/p(x_s)$ is large).

However, directly computing $\pi(x \mid y) = \pi(y \mid x)\,p(x) / q(y)$ requires the intractable response marginal $q(y)$.
Section~\ref{sec:role_switch} shows that for translation-invariant pairwise methods, $q(y)$ cancels entirely, making the optimization practical.

%%% ---- Section 3.5: Role-Switch ----
\subsection{Role-Switch: Fine-tuning $\pi(x \mid y)$ via Existing Frameworks}
\label{sec:role_switch}

The abductive policy follows from Bayes' theorem:
\begin{equation}
\label{eq:abductive_pi}
    \widetilde{\pi}(x \mid y) = \frac{\pi(y \mid x)\,p(x)}{q(y)},
\end{equation}
where $p(x)$ is the population prompt marginal and $q(y) = q_\pi(y)$ the induced response marginal; we write $\widetilde{\pi}(x \mid y)$, or $\pi(x \mid y)$ where unambiguous, for this induced conditional, with instances $\widetilde{\pi}_\theta$ and $\widetilde{\pi}_{\mathrm{ref}}$ for $\pi_\theta$ and $\pi_{\mathrm{ref}}$.
Direct training of $\widetilde{\pi}$ requires the intractable marginals $q(y)$ and $p(x)$; the following result bypasses both (proof in Appendix~\ref{app:prop}).

\begin{theorem}[Role-Switch Equivalence]
\label{prop:general}
Let $\mathcal{L} = -\sum_i F(\psi(x_{w,i}, y_i),\, \psi(x_{l,i}, y_i))$ with $\psi$ and $F$ as defined in Theorem~\ref{thm:forward_balanced}.
Suppose:
\begin{enumerate}[label=(\alph*)]
    \item the prompt marginal $p(x)$ is fixed, independent of $\pi_\theta$ and $\pi_{\mathrm{ref}}$;
    \item $F$ is translation-invariant: $F(a{+}c,\, b{+}c) = F(a, b)$ for all $c$.
\end{enumerate}
Then, for a dataset $\mathcal{D}$ of curated abductive triples $(x_w, x_l, y)$, the abductive loss $\widetilde{\mathcal{L}}$, defined as $\mathcal{L}$ with $\psi$ replaced by the abductive score $\widetilde{\psi}(x, y) := \log \widetilde{\pi}_\theta(x \mid y) - \log \widetilde{\pi}_{\mathrm{ref}}(x \mid y)$, reduces to the role-switched forward loss:
\begin{equation}
\label{eq:general_role_switch}
    \widetilde{\mathcal{L}}(\theta) = -\mathbb{E}_{(x_w, x_l, y)\sim\mathcal{D}}\!\left[F\!\left(\psi(x_w, y),\; \psi(x_l, y)\right)\right],
\end{equation}
requiring no computation of $\widetilde{\pi}$, $p(x)$, or $q(y)$.
\end{theorem}
Both conditions are satisfied by DPO, IPO, GPO, and SLiC-HF.
For DPO ($F(a,b) = \log\sigma(\beta(a-b))$, with $\sigma$ the logistic function and $\beta$ the inverse-temperature parameter), this yields the A-DPO loss:
\begin{equation}
    \label{eq:adpo}
        \mathcal{L}_{\mathrm{A\text{-}DPO}} = -
        \mathbb{E}_{(x_w, x_l, y)\sim\mathcal{D}}\!\left[ \log \sigma\big(\beta (\psi(x_w, y) - \psi(x_l, y))\big)\right].
\end{equation}

The role-switch extends beyond pairwise methods: for contrastive losses such as InfoNCE~\citep{oord2018representation} that compare against multiple negatives, $q(y)$ cancels through a different mechanism:

\begin{corollary}[Extension to Contrastive Losses]
\label{cor:infonce}
Fix a training-time sampling distribution $p_{\mathcal{S}}$ that draws each contrastive tuple $(x_1,\ldots,x_K,\,y)$ with uniform within-tuple candidates, $p_{\mathcal{S}}(x_j) = 1/K$ for all $j$, and $x^+ := x_1$ as the positive. Consider InfoNCE with score $s(x, y) = \log \pi_\theta(y \mid x)$. Then swapping the data roles yields exact optimization of the abductive conditional induced by the training joint $p_{\mathcal{S}}(x,y) := p_{\mathcal{S}}(x)\,\pi_\theta(y\mid x)$, namely
\begin{equation*}
    \pi_{\mathcal{S}}(x \mid y) \;:=\; \frac{p_{\mathcal{S}}(x)\,\pi_\theta(y\mid x)}{\sum_{x'} p_{\mathcal{S}}(x')\,\pi_\theta(y\mid x')}.
\end{equation*}
\end{corollary}
Proof in Appendix~\ref{app:infonce}.

Here $p_{\mathcal{S}}$ is set by the data pipeline: every training tuple in our datasets contains exactly one $x_w$ and one $x_l$, so within-tuple uniformity holds by construction. The target $\pi_{\mathcal{S}}(x \mid y)$ is the posterior under this uniform prompt prior and differs from $\widetilde{\pi}(x \mid y)$ by the prior log-ratio $\log p(x_w) - \log p(x_l)$.

For methods like DPOP~\citep{pal2024smaug} that include a penalty term $r(a) = -\lambda_{\mathrm{DPOP}} \max(0, -a)$ breaking strict translation invariance, the role-switch is not exact but the approximation error is bounded:

\begin{corollary}[Bounded Approximation for A-DPOP]
\label{cor:dpop_bound}
Write the DPOP loss as $F_{\mathrm{DPOP}}(a, b) = h(a - b) + r(a)$ where $h(t) = \log\sigma(\beta t)$.
Let $\widetilde{\mathcal{L}}_{\mathrm{DPOP}}$ denote the true abductive DPOP loss from pairwise $\pi(x \mid y)$ learning, and $\mathcal{L}_{\mathrm{A\text{-}DPOP}}$ the practical role-switched loss.
Then:
\begin{equation}
\label{eq:dpop_bound}
    \big|\widetilde{\mathcal{L}}_{\mathrm{DPOP}}(\theta) - \mathcal{L}_{\mathrm{A\text{-}DPOP}}(\theta)\big| \leq \lambda_{\mathrm{DPOP}} \cdot \mathbb{E}_{y}\!\left[|C(y)|\right],
\end{equation}
where $C(y) := \log(q_{\mathrm{ref}}(y) / q_\theta(y))$ is the log-ratio of the response marginals induced by $\pi_{\mathrm{ref}}$ and $\pi_\theta$.
At initialization $\theta = \theta_{\mathrm{ref}}$, $C(y) = 0$ and the bound is exactly zero; during fine-tuning, $\mathbb{E}[|C(y)|]$ remains small when parameter updates are limited.
\end{corollary}
Proof in Appendix~\ref{app:dpop_bound}.
Multi-DPOP achieves strong performance across all benchmarks (Section~\ref{sec:expanded_vlm}).

We call the family of role-switched methods \textbf{abductive preference learning}.

\subsection{Gradient Recovery Under Role-Switch}
At the gradient level, the role-switch resolves the starvation identified in Theorem~\ref{thm:starvation}.
% Under the conditions of Theorem~\ref{prop:general} (translation-invariant $F$; stated in Section~\ref{sec:role_switch}), the gradient of the role-switched loss with backward preference data $(x_w, x_l, y)$ where $x_w - x_l = v_d$ satisfies:

\begin{theorem}[Gradient Recovery of Detail Features]
\label{thm:recovery}
For any translation-invariant pairwise loss with backward data $(x_w, x_l, y)$ where $x_w - x_l = v_d$, the gradient decomposes as:
\begin{equation}
\label{eq:bwd_grad}
    \nabla_\theta \mathcal{L}_{\mathrm{bwd}} = -\sum_{i=1}^{N} F'(s_i) \cdot g(x_{l,i}, y_i) \cdot \mathbf{H}_\theta(x_{l,i}) \, v_d + O(\|v_d\|^2),
\end{equation}
where $f_\theta(x)$ is the prompt-feature map through which $x$ enters $\log \pi_\theta(y \mid x)$, $g(x, y) := \partial \log \pi_\theta(y \mid x) / \partial f_\theta(x)$ is the response sensitivity to that feature, and $\mathbf{H}_\theta(x) := \partial^2 f_\theta(x) / \partial \theta\, \partial x^{\top}$ is its parameter--input mixed Hessian, the single-output analogue of $\bar{\mathbf{H}}_\theta$ in Theorem~\ref{thm:starvation}.
The common-feature component $\nabla_\theta f_\theta(x_l)$ is exactly cancelled by the contrastive subtraction.
Every sample contributes detail signal $\mathbf{H}_\theta(x_l)\, v_d$, regardless of $N_g/N_s$.
\end{theorem}
Proof in Appendix~\ref{sec:gradient_recovery}.
In contrast to Theorem~\ref{thm:starvation}, where the detail component scales as $O(N_s)$ and vanishes under frequency imbalance, the backward detail component scales as $O(N)$ and is independent of the data distribution.

\section{Experimental Setup}

All datasets are converted to abductive preference triples $(x_w, x_l, y)$ via the role-switch of Theorem~\ref{prop:general}. Full construction details, dataset statistics, and abductive-pair examples are in Appendix~\ref{app:dataset_construction}.

\begin{figure*}[htbp]
\centering
\begin{minipage}[t]{0.60\linewidth}
\centering
\footnotesize
\resizebox{\linewidth}{!}{
\begin{tabular}{m{0.78\linewidth} m{0.2\linewidth}}
\toprule
\multicolumn{1}{c}{\textbf{Prompts}} & \multicolumn{1}{c}{\textbf{Answer}} \\
\midrule
{\bf Background Knowledge:} Yannis Philippakis (born 23 April 1986) is the lead singer and guitarist of the British indie rock band Foals. Dorothee Pesch (born 3 June 1964), popularly known as Doro Pesch or Doro, is a [\textcolor{deepblue}{\bf German} / \textcolor{deepred}{\bf Canadian}] heavy metal singer-songwriter, formerly front-woman of the heavy metal band Warlock.

{\bf Question:} Of the two artists, Yannis Philippakis and Dorothee Pesch, whose country of origin is geographically closer to Austria?
&
Yannis Philippakis is closer to Austria. \\
\bottomrule
\end{tabular}
}
\vspace{4pt}
\captionof{table}{\textsc{A-HaluEval} construction: minimal background-knowledge edits. \textcolor{deepblue}{\bf Blue}: original; \textcolor{deepred}{\bf red}: edited.}
\label{tab:abductive_halueval}
\end{minipage}%
\hfill
\begin{minipage}[t]{0.36\linewidth}
\centering
\includegraphics[width=\linewidth]{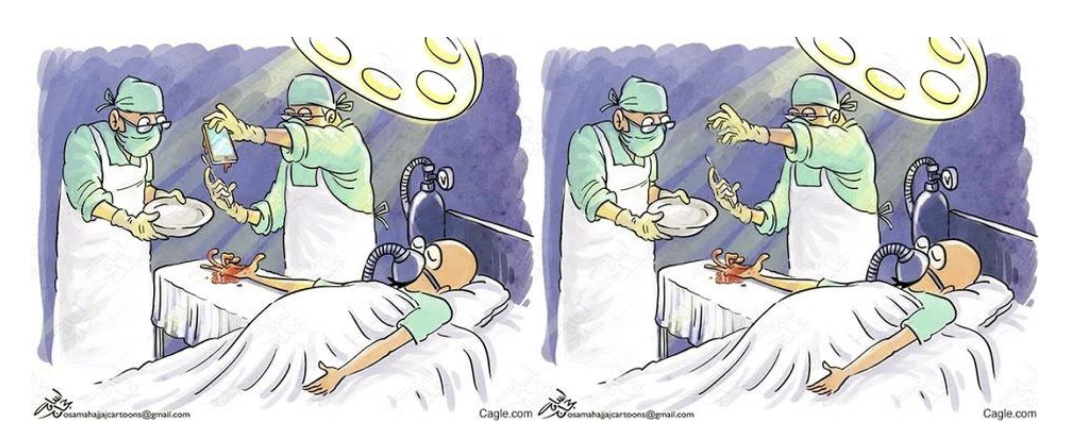}
\vspace{4pt}
\caption{Contrastive pair from \textsc{HumorDB}. Left: funny ($83.3\%$). Right: not funny ($85.7\%$). The phone in the surgeon's hand drives the humor.}
\label{fig:a-halueval}
\end{minipage}
\end{figure*}

\paragraph{Datasets.}
\textit{Multimodal:}
(i)~\textsc{HumorDB}~\citep{jain2024ai} — contrastive image pairs where subtle edits flip humor perception ($991$ train, $300$ test; Figure~\ref{fig:a-halueval});
(ii)~\textsc{VLMBias}~\citep{vo2025visionlanguagemodelsbiased} — counterfactual images that break visual priors, e.g., a dog with 5 legs ($847$ train, $200$ test).
\textit{Text:}
(iii)~\textsc{HaluEval}~\citep{li2023halueval} — factual QA where minimal background-knowledge edits flip the correct answer (Table~\ref{tab:abductive_halueval});
(iv)~\textsc{Inverse-IFEval}~\citep{zhang2025inverseifevalllmsunlearn} — counterintuitive constraint reformulations of IFEval-style instructions~\citep{zhou2023instructionfollowingevaluationlargelanguage}, testing whether models can break from learned instruction-following patterns.

\paragraph{Models.}
\textit{VLM (5 models):} \textsc{Qwen2.5-VL-7B}~\citep{bai2025qwen25vltechnicalreport}, \textsc{Qwen3-VL-4B/8B}~\citep{bai2025qwen3vltechnicalreport}, \textsc{InternVL3.5-8B}~\citep{wang2025internvl3_5}, and \textsc{Molmo2-4B}~\citep{clark2026molmo2openweightsdata}. We additionally evaluate six closed-source VLMs in zero-shot: \textsc{GPT-5.2}, \textsc{Qwen3-VL-235B}, \textsc{Gemini 3 Flash}, \textsc{Kimi-K2.5}, and \textsc{Claude Sonnet 4.5/4.6}.
\textit{Text (5 LLMs):} \textsc{Tulu-3.1-8B}~\citep{lambert2025tulu3pushingfrontiers}, \textsc{Gemma-3-4B-it}~\citep{gemmateam2025gemma3technicalreport}, \textsc{GLM-4-9B-0414}~\citep{glm2024chatglm}, \textsc{Qwen3-4B}~\citep{yang2025qwen3technicalreport}, and \textsc{Qwen3-30B-A3B}~\citep{yang2025qwen3technicalreport}.

\paragraph{Baselines and training.}
We compare against three baseline paradigms: (i)~SFT on the preferred completions $y_w$ only (no rejected responses), which may affect general capability since training is restricted to curated data; (ii)~DPO/DPOP~\citep{rafailov2023direct,pal2024smaug} as offline preference baselines; and (iii)~GRPO~\citep{shao2024deepseekmathpushinglimitsmathematical} as an online RL baseline with binary correctness reward (RLVR).
Our methods are A-DPO (\Cref{eq:adpo}), Multi-DPO (\Cref{eq:multi}), and their DPOP variants (Corollary~\ref{cor:dpop_bound}), with $\beta{=}0.1$ and $\lambda \in \{0, 0.5, 1\}$. For VLMs, we use A-DPO with full fine-tuning ($\beta{=}0.1$). Hyperparameters are in Appendix~\ref{app:hyperparameters}; training details per dataset are in Appendix~\ref{app:training_details}.

\paragraph{Multitask objective.}
Since forward and abductive losses address complementary axes (response selection vs.\ prompt discrimination), we combine them:
\begin{equation}
    \label{eq:multi}
        \mathcal{L}_{\text{Multi-DPO}}(\lambda) := \lambda \,\mathcal{L}_{\text{DPO}} + (1-\lambda)\, \mathcal{L}_{\text{A-DPO}},
\end{equation}
where $\mathcal{L}_{\text{DPO}}$ is the standard forward DPO loss on $(x, y_w, y_l)$ triples and $\lambda \in (0,1)$; we use $\lambda = 0.5$ by default. The same construction yields $\mathcal{L}_{\text{Multi-DPOP}}$.

\section{Results}
\label{sec:exp}

We evaluate using pairwise log-probability comparison for \textsc{HaluEval}/\textsc{A-HaluEval}, and generation-based accuracy for \textsc{Inverse-IFEval} (deterministic rule-based parsing), \textsc{HumorDB}, and \textsc{VLMBias}. General capability is measured via MMLU-Pro, GPQA-D, AlpacaEval-2, SimpleQA, TruthfulQA (LLMs) and MMMU-Pro, HallusionBench (VLMs). All results are averaged over 3 runs ($\pm$ std); full definitions in Appendix~\ref{app:metrics}.

\subsection{RQ1: Does A-DPO Recover Long-Tail Prompt Sensitivity?}
\label{sec:expanded_vlm}

\paragraph{Vision-language models.}
Table~\ref{tab:vlm_comprehensive} shows that base VLMs score near-chance on both \textsc{HumorDB} ($38$--$45\%$) and \textsc{VLMBias} ($2$--$3\%$). Forward baselines progressively improve HumorDB (GRPO to $58$--$61\%$) and modestly increase VLMBias (GRPO to $8$--$12\%$), but fall far short of A-DPO. By reversing the conditioning direction, A-DPO achieves $83$--$85\%$ on HumorDB and $22$--$44\%$ on VLMBias — with Qwen3-VL-8B reaching $43.5\%$, a $14\times$ improvement over its $3.1\%$ base. Most closed-source models (Qwen3-VL-235B at $0\%$, Claude Sonnet 4.6 at $0\%$, GPT-5.2 at $4.6\%$) cannot overcome visual prior bias in zero-shot, with Gemini 3 Flash ($67.7\%$) as the sole exception — likely due to its omni-model architecture trained end-to-end rather than via a frozen vision encoder~\citep{gemmateam2025gemma3technicalreport}. Our post-training approach is complementary and architecture-agnostic.

\begin{table}[htbp]
    \centering
    \caption{Multimodal results on \textsc{HumorDB} and \textsc{VLMBias}. Forward baselines improve HumorDB incrementally but barely shift VLMBias; A-DPO excels on both. Closed-source models included for reference.}
    \vspace{2pt}
    \label{tab:vlm_comprehensive}
    \resizebox{0.9\linewidth}{!}{
    \begin{tabular}{lcccccccccc}
    \toprule
    & \multicolumn{5}{c}{\textbf{HumorDB}} & \multicolumn{5}{c}{\textbf{VLMBias}} \\
    \cmidrule(lr){2-6} \cmidrule(lr){7-11}
    \textbf{Model} & Base & SFT & DPO & GRPO & A-DPO & Base & SFT & DPO & GRPO & A-DPO \\
    \midrule
    \multicolumn{11}{c}{\textit{Closed-Source Models (Zero-Shot)}} \\
    \midrule
    GPT-5.2              & $87.0$          & -- & -- & --  & --  & $4.6$           & -- & -- & --  & -- \\
    Qwen3-VL-235B        & $\mathbf{90.0}$ & -- & -- & --  & --  & $0.0$           & -- & -- & --  & -- \\
    Gemini 3 Flash       & $86.0$          & -- & -- & --  & --  & $\mathbf{67.7}$ & -- & -- & --  & -- \\
    Kimi-K2.5            & $61.0$          & -- & -- & --  & --  & $1.5$           & -- & -- & --  & -- \\
    Claude Sonnet 4.6    & $59.0$          & -- & -- & --  & --  & $0.0$           & -- & -- & --  & -- \\
    Claude Sonnet 4.5    & $51.0$          & -- & -- & --  & --  & $0.0$           & -- & -- & --  & -- \\
    \midrule
    \multicolumn{11}{c}{\textit{Open-Source Models}} \\
    \midrule
    Qwen3-VL-4B    & $38.3$ & $48.5_{\pm 0.9}$ & $53.7_{\pm 1.7}$ & $58.2_{\pm 1.1}$ & $82.5_{\pm 0.6}$          & $2.3$ & $3.1_{\pm 0.8}$ & $5.8_{\pm 1.3}$ & $9.4_{\pm 1.7}$ & $28.7_{\pm 2.3}$ \\
    Qwen3-VL-8B    & $44.7$ & $53.2_{\pm 1.6}$ & $57.1_{\pm 0.7}$ & $61.4_{\pm 1.3}$ & $\mathbf{85.0_{\pm 0.5}}$ & $3.1$ & $4.2_{\pm 1.1}$ & $7.3_{\pm 1.5}$ & $11.8_{\pm 2.0}$ & $\mathbf{43.5_{\pm 2.1}}$ \\
    Qwen2.5-VL-7B  & $40.7$ & $50.8_{\pm 1.1}$ & $55.4_{\pm 1.5}$ & $59.7_{\pm 0.8}$ & $83.5_{\pm 0.7}$          & $2.5$ & $3.5_{\pm 0.9}$ & $6.1_{\pm 1.4}$ & $10.2_{\pm 1.8}$ & $31.2_{\pm 2.5}$ \\
    InternVL3-8B   & $42.0$ & $51.4_{\pm 1.5}$ & $56.2_{\pm 1.0}$ & $60.3_{\pm 1.6}$ & $83.5_{\pm 0.8}$          & $2.8$ & $3.8_{\pm 1.0}$ & $6.5_{\pm 1.2}$ & $10.6_{\pm 1.6}$ & $29.8_{\pm 2.8}$ \\
    Molmo2-4B      & $42.3$ & $49.1_{\pm 0.7}$ & $53.5_{\pm 1.8}$ & $57.8_{\pm 1.2}$ & $83.0_{\pm 1.1}$          & $2.1$ & $2.9_{\pm 0.6}$ & $4.7_{\pm 1.1}$ & $8.1_{\pm 1.9}$ & $22.4_{\pm 2.4}$ \\
    \bottomrule
    \end{tabular}
    }
    \end{table}
    
\paragraph{Language models.}
Table~\ref{tab:additional_summary} confirms the same pattern across five LLM families. SFT, DPO, and GRPO all improve standard HaluEval accuracy but leave abductive accuracy near base levels ($54$--$61\%$). Multi-DPOP achieves near-perfect standard accuracy ($98.5$--$100\%$) while reaching $82.5$--$91.5\%$ abductive accuracy — a $25$--$35\%$ absolute gain. On \textsc{Inverse-IFEval}, Multi-DPOP reaches $65$--$84\%$, surpassing Claude-4 Opus ($67.2\%$) even at 4B scale and exceeding GPT-5-high ($73.7\%$) at 9B+. DPO degrades IFBench ($-8$--$12\%$); A-DPO and Multi-DPOP preserve it ($\pm 0.5\%$).

\label{sec::main}

\begin{table}[htbp]
\centering
\caption{LLM results across five model families ($\pm$ std over 3 runs). Multi-DPOP surpasses Claude-4-Opus ($67.2\%$) and approaches GPT-5 ($73.7\%$) on Inverse-IFEval even at 4B--9B scale, while achieving $82$--$92\%$ abductive accuracy. Full per-method breakdown in Appendix Table~\ref{tab:additional_summary_full}.}
\vspace{2pt}
\label{tab:additional_summary}
\resizebox{\linewidth}{!}{%
\begin{tabular}{lccc ccc ccc}
\toprule
& \multicolumn{3}{c}{\textbf{HaluEval Acc.}} & \multicolumn{3}{c}{\textbf{A-HaluEval (Abductive)}} & \multicolumn{3}{c}{\textbf{Inverse-IFEval}} \\
\cmidrule(lr){2-4} \cmidrule(lr){5-7} \cmidrule(lr){8-10}
\textbf{Model} & Base & GRPO & M-DPOP & Base & GRPO & M-DPOP & Base & GRPO & M-DPOP \\
\midrule
\multicolumn{10}{c}{\textit{Closed-Source (Zero-Shot, Inv-IF only)}} \\
\midrule
Claude-4.5-Sonnet  & -- & -- & -- & -- & -- & -- & $67.2$ & -- & -- \\
Gemini-3-Flash & -- & -- & -- & -- & -- & -- & $70.6$ & -- & -- \\
GPT-5.2     & -- & -- & -- & -- & -- & -- & $73.7$ & --- & -- \\
\midrule
\multicolumn{10}{c}{\textit{Open-Source (Ours)}} \\
\midrule
Tulu-3.1-8B  & $90.0$ & $97.5_{\pm 0.3}$ & $\mathbf{99.5_{\pm 0.2}}$ & $54.7$ & $58.3_{\pm 0.9}$ & $\mathbf{85.0_{\pm 1.4}}$ & $32.8$ & $40.2_{\pm 1.8}$ & $\mathbf{65.3_{\pm 1.6}}$ \\
Gemma-3-4B   & $87.5$ & $95.8_{\pm 0.4}$ & $\mathbf{98.5_{\pm 0.3}}$ & $52.3$ & $56.1_{\pm 1.1}$ & $\mathbf{82.5_{\pm 0.8}}$ & $38.5$ & $43.7_{\pm 1.3}$ & $\mathbf{69.8_{\pm 1.9}}$ \\
GLM-4-9B     & $88.3$ & $96.2_{\pm 0.6}$ & $\mathbf{99.0_{\pm 0.1}}$ & $53.8$ & $57.4_{\pm 1.6}$ & $\mathbf{84.0_{\pm 1.2}}$ & $42.1$ & $48.5_{\pm 1.2}$ & $\mathbf{74.2_{\pm 1.7}}$ \\
Qwen3-4B     & $89.1$ & $97.2_{\pm 0.3}$ & $\mathbf{99.5_{\pm 0.2}}$ & $55.2$ & $59.3_{\pm 1.7}$ & $\mathbf{86.5_{\pm 0.6}}$ & $44.7$ & $51.2_{\pm 1.6}$ & $\mathbf{76.1_{\pm 1.5}}$ \\
Qwen3-30B    & $91.2$ & $98.1_{\pm 0.2}$ & $\mathbf{100.0_{\pm 0.0}}$ & $57.6$ & $61.4_{\pm 0.7}$ & $\mathbf{91.5_{\pm 0.9}}$ & $49.2$ & $56.8_{\pm 0.9}$ & $\mathbf{83.7_{\pm 1.3}}$ \\
\bottomrule
\end{tabular}
}
\end{table}

\subsection{RQ2: Why Do Forward Methods Fail?}
\label{sec:inverse_ifeval_results}

The failure of SFT, DPO, and GRPO on the abductive axis is structural, not a capacity issue.
All forward baselines leave abductive accuracy near-unchanged ($54$--$61\%$ A-HaluEval; Table~\ref{tab:additional_summary}). On Inverse-IFEval, forward methods improve modestly ($38$--$57\%$ vs.\ $33$--$49\%$ base) while DPO \emph{significantly degrades} IFBench ($-8$--$12\%$; Appendix Table~\ref{tab:inverse_ifeval}), showing that standard preference training overfits to constraint patterns at the expense of general instruction following. Multi-DPOP reaches $65$--$84\%$, surpassing frontier closed-source models, while fully preserving IFBench ($\pm 0.5\%$). This is consistent with Theorem~\ref{thm:starvation}: forward training starves the prompt-axis gradient regardless of paradigm.

\subsection{RQ3: Are General Capabilities Preserved?}
\label{sec:general_capability}

Table~\ref{tab:generation_quality} reports five general capability benchmarks for the text models Tulu-3.1-8B and Qwen3-30B. Performance on MMLU-Pro, GPQA-D, AlpacaEval-2, and SimpleQA changes by less than $1\%$ across DPO, A-DPO, and Multi-DPOP. SFT decreases AlpacaEval scores by $1.8\%$ and MMLU-Pro scores by $0.6\%$. A-DPO increases TruthfulQA scores by $2.6\%$ on Tulu-3.1-8B and $2.5\%$ on Qwen3-30B. This increase supports the hypothesis that abductive training reduces overconfident factual priors.

For vision-language models, Table~\ref{tab:vlm_general} shows performance before and after A-DPO. MMMU-Pro scores change by less than $0.6\%$. \textsc{HallusionBench} scores increase by $5\%$ to $6\%$ absolute.

\begin{table}[t]
\centering
\caption{General capability benchmarks for text models. Results are averaged over 3 runs ($\pm$ std). All evaluations use the unified pipeline (Appendix~\ref{app:hyperparameters}).}
\vspace{2pt}
\label{tab:generation_quality}
\resizebox{\linewidth}{!}{
\begin{tabular}{lcccccccccc}
\toprule
& \multicolumn{5}{c}{\textbf{Tulu-3.1-8B}} & \multicolumn{5}{c}{\textbf{Qwen3-30B-A3B}} \\
\cmidrule(lr){2-6} \cmidrule(lr){7-11}
\textbf{Method} & MMLU-Pro & GPQA-D & AlpacaEval & SimpleQA & TruthfulQA & MMLU-Pro & GPQA-D & AlpacaEval & SimpleQA & TruthfulQA \\
\midrule
Base Model & $20.2_{\pm 0.3}$ & $6.3_{\pm 0.8}$ & $34.5_{\pm 0.5}$ & $5.1_{\pm 0.4}$ & $59.9_{\pm 0.6}$ & $61.5_{\pm 0.2}$ & $44.0_{\pm 0.7}$ & $67.4_{\pm 0.4}$ & $18.3_{\pm 0.3}$ & $62.8_{\pm 0.5}$ \\
\midrule
SFT        & $19.6_{\pm 0.4}$ & $5.8_{\pm 1.1}$ & $32.8_{\pm 0.6}$ & $4.8_{\pm 0.5}$ & $59.1_{\pm 0.7}$ & $60.7_{\pm 0.3}$ & $43.1_{\pm 0.9}$ & $65.6_{\pm 0.5}$ & $17.5_{\pm 0.4}$ & $62.0_{\pm 0.6}$ \\
DPO        & $20.0_{\pm 0.3}$ & $6.1_{\pm 0.9}$ & $36.7_{\pm 0.4}$ & $5.0_{\pm 0.4}$ & $59.6_{\pm 0.5}$ & $61.3_{\pm 0.2}$ & $43.8_{\pm 0.8}$ & $69.2_{\pm 0.3}$ & $18.1_{\pm 0.3}$ & $62.5_{\pm 0.5}$ \\
A-DPO      & $20.1_{\pm 0.3}$ & $6.4_{\pm 0.9}$ & $34.2_{\pm 0.5}$ & $5.2_{\pm 0.4}$ & $\mathbf{62.5_{\pm 0.6}}$ & $61.4_{\pm 0.2}$ & $43.9_{\pm 0.8}$ & $67.0_{\pm 0.4}$ & $18.4_{\pm 0.3}$ & $\mathbf{65.3_{\pm 0.5}}$ \\
Multi-DPO  & $20.1_{\pm 0.3}$ & $5.9_{\pm 1.0}$ & $35.4_{\pm 0.4}$ & $5.0_{\pm 0.4}$ & $61.5_{\pm 0.6}$ & $61.4_{\pm 0.2}$ & $44.2_{\pm 0.7}$ & $68.1_{\pm 0.3}$ & $18.2_{\pm 0.3}$ & $64.1_{\pm 0.5}$ \\
\midrule
Multi-DPOP & $\mathbf{20.3_{\pm 0.3}}$ & $6.2_{\pm 0.8}$ & $\mathbf{36.9_{\pm 0.4}}$ & $\mathbf{5.3_{\pm 0.4}}$ & $61.9_{\pm 0.6}$ & $\mathbf{61.6_{\pm 0.2}}$ & $\mathbf{44.3_{\pm 0.7}}$ & $\mathbf{69.7_{\pm 0.3}}$ & $\mathbf{18.5_{\pm 0.3}}$ & $64.9_{\pm 0.5}$ \\
\bottomrule
\end{tabular}
}
\end{table}

\begin{table}[htbp]
\centering
\caption{Vision-language model performance before and after A-DPO.}
\vspace{2pt}
\label{tab:vlm_general}
\resizebox{0.6\linewidth}{!}{%
\begin{tabular}{lcccc}
\toprule
& \multicolumn{2}{c}{\textbf{MMMU-Pro}} & \multicolumn{2}{c}{\textbf{HallusionBench}} \\
\cmidrule(lr){2-3} \cmidrule(lr){4-5}
\textbf{Model} & Base & A-DPO & Base & A-DPO \\
\midrule
Qwen3-VL-8B    & $55.9_{\pm 0.4}$ & $56.3_{\pm 0.5}$ & $61.1_{\pm 0.8}$ & $\mathbf{66.8_{\pm 0.9}}$ \\
InternVL3-8B   & $43.7_{\pm 0.4}$ & $44.2_{\pm 0.5}$ & $49.9_{\pm 0.9}$ & $55.3_{\pm 1.0}$ \\
Qwen2.5-VL-7B  & $41.0_{\pm 0.3}$ & $41.5_{\pm 0.4}$ & $52.9_{\pm 0.8}$ & $58.1_{\pm 0.9}$ \\
\bottomrule
\end{tabular}
}
\end{table}

\subsection{RQ4: Does the Role-Switch Generalize Across Methods?}
\label{sec::textonly}

\begin{wraptable}{r}{0.38\linewidth}
\vspace{-14pt}
\centering
\caption{Ablations on VLMBias (Qwen3-VL-8B).}
\vspace{2pt}
\label{tab:aipo}
\resizebox{\linewidth}{!}{%
\begin{tabular}{lcc}
\toprule
\textbf{Method} & \textbf{HumorDB} & \textbf{VLMBias} \\
\midrule
Base       & $44.7$ & $3.1$ \\
\midrule
DPO        & $57.1$ & $7.3$ \\
CDA-DPO    & $59.8$ & $12.4$ \\
IPO        & $55.8$ & $6.8$ \\
\midrule
A-DPO      & $\mathbf{85.0}$ & $\mathbf{43.5}$ \\
A-IPO      & $83.7$ & $40.1$ \\
\bottomrule
\end{tabular}
}
\vspace{-25pt}
\end{wraptable}

Our theory (Theorem~\ref{prop:general}) predicts that any translation-invariant pairwise method benefits from the role-switch. We validate this along three axes (Table~\ref{tab:aipo}).

\paragraph{Role-switch vs.\ data augmentation.}
CDA-DPO uses the \emph{same} $(x_w, x_l, y)$ triples as A-DPO but in standard forward DPO format. It achieves only $12.4\%$ on VLMBias vs.\ A-DPO's $43.5\%$, confirming the reversed comparison axis instead of data exposure, which drives the gain.

\paragraph{Generalization across losses.}
A-IPO~\citep{azar2024general} (squared loss) achieves $40.1\%$ on VLMBias, matching A-DPO's $43.5\%$ (log-sigmoid), confirming the role-switch is loss-agnostic.

Additional ablations: $\lambda$ sensitivity (stable for $[0.2, 0.7]$; Figure~\ref{fig:lambda_ablation}), data efficiency (25\% of data already $2\times$ above GRPO; Table~\ref{tab:data_ablation}), $\beta$ sensitivity, and DPOP penalty are in Appendix~\ref{app:additional_ablations}.

\section{Conclusion}
We identified \emph{long-tail prompt sensitivity} as a structural failure mode of preference learning, all forward methods compare responses given a fixed prompt, leaving the prompt axis unoptimized, and introduced \textbf{abductive preference learning}, which reverses the comparison axis.
We proved that this reversal recovers the starved detail-feature gradient for any translation-invariant pairwise method via a simple data role-switch, requiring no architectural changes.
Across four benchmarks and two modalities, abductive training consistently outperforms SFT, DPO, and GRPO: A-DPO raises VLMBias accuracy from $3\%$ to $44\%$, surpassing GPT-5.2 ($4.6\%$); Multi-DPOP reaches $65$--$84\%$ on Inverse-IFEval, exceeding GPT-5 ($73.7\%$) at the 9B scale while preserving IFBench; and similar gains hold on HumorDB ($85\%$) and A-HaluEval ($92\%$ abductive accuracy), all without degrading general capabilities across five LLM and five VLM families.
These findings suggest that learning $\pi(x \mid y)$: asking ``which prompt justifies this response?'', is a broadly useful and complementary signal to the standard $\pi(y \mid x)$ objective, applicable to any existing preference learning pipeline at zero additional cost.

\section{Impact Statement}
Abductive preference learning improves model sensitivity to fine-grained input variations, which has positive implications for safety-critical applications (e.g., medical QA where a single changed detail alters the correct answer).
However, the same capability could be misused to train models that are more sensitive to adversarial prompt perturbations.
We believe the benefits outweigh the risks, as the method primarily corrects an existing failure mode (insensitivity to relevant details) rather than introducing new capabilities.

\bibliography{refs}
\bibliographystyle{colm2026_conference}

%%%%%%%%%%%%%%%%%%%%%%%%%%%%%%%%%%%%%%%%%%%%%%%%%%%%%%%%%%%%%%%%%%%%%%%%%%%%%%%
%%%%%%%%%%%%%%%%%%%%%%%%%%%%%%%%%%%%%%%%%%%%%%%%%%%%%%%%%%%%%%%%%%%%%%%%%%%%%%%
% APPENDIX
%%%%%%%%%%%%%%%%%%%%%%%%%%%%%%%%%%%%%%%%%%%%%%%%%%%%%%%%%%%%%%%%%%%%%%%%%%%%%%%
%%%%%%%%%%%%%%%%%%%%%%%%%%%%%%%%%%%%%%%%%%%%%%%%%%%%%%%%%%%%%%%%%%%%%%%%%%%%%%%
\newpage
\appendix
\section{Limitations}
\begin{enumerate}[leftmargin=*]
    \item \textbf{Dataset scale.} Our text abductive dataset (\textsc{A-HaluEval}) contains 1,001 entries. While effective at this scale, behavior at 10K--100K entries is untested. The method's effectiveness on externally curated datasets (\textsc{HumorDB}, \textsc{VLMBias}, \textsc{Inverse-IFEval}) suggests the gains are not specific to our construction pipeline.

    \item \textbf{First-order approximation.} Theorems~\ref{thm:starvation}--\ref{thm:recovery} are stated for general twice-differentiable models but rely on a first-order Taylor expansion ($O(\|v_d\|^2)$ remainder). While the predictions are consistent with empirical results in full-scale transformers, a tighter analysis bounding the higher-order terms for specific architectures (e.g., attention layers) would strengthen the theoretical foundation.

    \item \textbf{Abductive data construction.} Constructing $(x_w, x_l, y)$ triples requires knowing how to minimally edit prompts in a semantically meaningful way. For text, this can be automated; for images, it requires curated minimally contrastive pairs (as in \textsc{VLMBias} and \textsc{HumorDB}). Scaling to arbitrary image domains remains open.

    \item \textbf{Multitask weight $\lambda$.} While $\lambda = 0.5$ works well across our experiments, the optimal balance may vary across domains and model scales. We provide a full ablation but no principled selection criterion.
\end{enumerate}

\section{Future Work}
Promising directions include: (i)~scaling abductive training to larger preference datasets with automated generation of minimally edited prompts; (ii)~tightening the first-order Taylor bounds for specific architectures (e.g., attention layers); (iii)~applying abductive preference learning to other modalities such as audio and video; and (iv)~combining abductive learning with online DPO to generate minimally edited prompts on-the-fly during training.
\section{Additional Results}
\label{app:additional_models}

\subsection{Dataset Construction}
\label{app:dataset_construction}
This appendix summarizes the construction of our abductive preference datasets.

\subsubsection{\textsc{A-HaluEval}: Construction and Validation}
For \textsc{HaluEval} QA, each example provides background knowledge, a question, and a hallucinated answer. We form $x_l$ by concatenating the original background knowledge and question, and form $x_w$ by applying a controlled edit to the background knowledge while keeping the question fixed; we use the original hallucinated answer as the shared response $y$ (Table~\ref{tab:abductive_halueval}).

Given a prompt-response pair $(x, y)$, denote the average log-likelihood of model $\pi$ corresponding to this pair as $\operatorname{ALL}_\pi(x, y)$. To ensure our data generation pipeline is reasonable, we employ a three-stage validation framework:
\begin{enumerate}[leftmargin=*]
    \item \textbf{Hallucination Verification:} Confirm that the base model produces hallucinated responses under the original background via enforcing an average log likelihood $\text{threshold}$.
    \item \textbf{Probability-Based Quality Assurance:} Verify that the hallucinated response is more likely under the modified background than the original:
\begin{equation}
    \label{eq:delta}
    \begin{split}
        & \operatorname{ALL}_\pi(\text{hallucination} \mid \text{original}) \\
        & \quad - \operatorname{ALL}_\pi(\text{hallucination} \mid \text{modified}) \geq \delta,
    \end{split}
\end{equation}
    where $\delta$ is a self-defined likelihood margin.
    \item \textbf{Contextual Reasonableness:} Ensure that the modified background logically supports the hallucinated answer through the validation of LLM agents.
\end{enumerate}

Notably, the parameter $\delta$ is set as $0.1$ without specification. Based on the fact that the \textsc{A-HaluEval} dataset is only equipped with $1,001$ entries compared with the $10,000$ entries in the original \textsc{HaluEval} dataset, we filtered out the original entries that lead to the generation of \textsc{A-HaluEval} for the DPO or DPOP training and the following Multi-DPO or Multi-DPOP training.

\subsubsection{\textsc{Inverse-IFEval}: Construction}
\label{app:inverse_ifeval_construction}

We construct abductive preference pairs from the \textsc{Inverse-IFEval} benchmark~\citep{zhang2025inverseifevalllmsunlearn}, which contains prompts with counterintuitive constraints that make the expected response seem wrong without the constraint. Our data generation pipeline proceeds as follows:

\paragraph{Stage 1: Base Model Filtering.}
We first filter entries where the base model fails to follow the constraint (judge score $= 0$). This ensures we only train on cases where the model genuinely struggles with counterintuitive instructions.

\paragraph{Stage 2: Constraint Extraction.}
Using GPT-4o as an agent, we extract the constraint and base question from each prompt, producing $(x_{\text{base}}, x_{\text{constrained}})$ pairs where the constraint modifies expected behavior.

\paragraph{Stage 3: Parallel Sampling.}
We generate $8$ candidate responses per entry using the base model with temperature $0.6$, then filter samples that fail the constraint-following check.

\paragraph{Stage 4: Base Question Validation (ADPO).}
For ADPO pairs, we validate that each failed response $y$ is actually reasonable for the base question $x_{\text{base}}$ using an LLM-as-judge. This ensures the ADPO formulation is correct: $(x_{\text{base}}, y)$ as chosen and $(x_{\text{constrained}}, y)$ as rejected.

\paragraph{Stage 5: DPO Pair Formation.}
For DPO pairs, we use the same constrained prompt with correct vs.\ incorrect responses: $(x_{\text{constrained}}, y_{\text{correct}})$ as chosen and $(x_{\text{constrained}}, y_{\text{wrong}})$ as rejected.

\paragraph{Dataset Statistics.}
Table~\ref{tab:inverse_ifeval_stats} summarizes the data generation results across five base models. We report the number of processed entries (out of $506$ total in the English split), ADPO and DPO training pairs formed, and base model accuracy on the constrained prompts.

\begin{table}[htbp]
\centering
\caption{Dataset generation statistics for \textsc{Inverse-IFEval}. Lower base model accuracy correlates with more training pairs, as models that struggle more provide more learning signal.}
\vspace{2pt}
\label{tab:inverse_ifeval_stats}
\resizebox{0.9\linewidth}{!}{
\begin{tabular}{lccccc}
\toprule
\textbf{Model} & \textbf{Processed} & \textbf{ADPO Pairs} & \textbf{DPO Pairs} & \textbf{Accuracy (\%)} \\
\midrule
Llama-3.1-Tulu-3.1-8B & 357/506 & 251 & 220 & 16.4 \\
Gemma-3-4B-it & 311/506 & 200 & 195 & 23.1 \\
GLM-4-9B-0414 & 318/506 & 211 & 225 & 27.5 \\
Qwen3-4B-Instruct & 303/506 & 193 & 201 & 29.1 \\
Qwen3-30B-A3B-Instruct & 324/506 & 239 & 197 & 31.4 \\
\bottomrule
\end{tabular}
}
\end{table}

\paragraph{Failure Analysis.}
Table~\ref{tab:inverse_ifeval_failures} breaks down why some entries could not form training pairs. The most common reasons are: (1) \textit{all\_samples\_succeeded}: the model correctly follows the constraint across all $8$ samples, providing no learning signal; (2) \textit{no\_valid\_base\_question}: no sample was valid for the base question while also failing the constraint; (3) \textit{constraint\_extraction\_failed}: the LLM agent could not reliably extract the constraint from the prompt.

\begin{table}[htbp]
\centering
\caption{Failure breakdown for \textsc{Inverse-IFEval} data generation. Better models (higher accuracy) have more ``all samples succeeded'' failures.}
\vspace{2pt}
\label{tab:inverse_ifeval_failures}
\resizebox{\linewidth}{!}{
\begin{tabular}{lcccc}
\toprule
\textbf{Model} & \textbf{All Succeeded} & \textbf{No Valid Base} & \textbf{Extract Failed} & \textbf{Validation Failed} \\
\midrule
Llama-3.1-Tulu-3.1-8B & 40 & 55 & 38 & 16 \\
Gemma-3-4B-it & 74 & 71 & 35 & 15 \\
GLM-4-9B-0414 & 71 & 69 & 31 & 17 \\
Qwen3-4B-Instruct & 96 & 61 & 33 & 13 \\
Qwen3-30B-A3B-Instruct & 110 & 34 & 24 & 10 \\
\bottomrule
\end{tabular}
}
\end{table}

This appendix provides full per-method results, detailed ablation studies, baseline method descriptions, theoretical proofs, and training hyperparameters.

\subsection{Baseline Methods}
\label{app:baselines}

We compare against three categories of forward-training baselines:

\paragraph{Supervised Fine-Tuning (SFT).}
SFT trains on the preferred completion $y_w$ via next-token prediction: $\mathcal{L}_{\text{SFT}} = -\mathbb{E}_{(x, y_w)} [\log \pi_\theta(y_w \mid x)]$. It does not use rejected responses and therefore has no contrastive signal. Because training is restricted to curated preferred completions (which may not be representative of the general data distribution), SFT can slightly degrade general capabilities (Table~\ref{tab:generation_quality}).

\paragraph{Direct Preference Optimization (DPO).}
DPO~\citep{rafailov2023direct} trains on preference pairs $(y_w, y_l \mid x)$ using the log-sigmoid loss over implicit reward margins: $\mathcal{L}_{\text{DPO}} = -\mathbb{E}[\log \sigma(\beta(\psi(x, y_w) - \psi(x, y_l)))]$, where $\psi(x, y) = \log \frac{\pi_\theta(y|x)}{\pi_{\text{ref}}(y|x)}$. DPOP~\citep{pal2024smaug} adds a penalty term to prevent chosen log-probability degradation. Both compare \emph{responses} given a fixed prompt, leaving the prompt axis unoptimized.

\paragraph{Group Relative Policy Optimization (GRPO).}
GRPO~\citep{shao2024deepseekmathpushinglimitsmathematical} is an online RL method that samples multiple completions per prompt, scores them with a reward function, and uses group-relative advantages for policy optimization. We use binary correctness as the reward signal: $r(x, y) = 1$ if the response is correct, $0$ otherwise. As a reinforcement learning from verifiable rewards (RLVR) approach, GRPO is strictly more powerful than DPO in that it generates on-policy data. Despite this, GRPO still optimizes the forward conditional $\pi(y \mid x)$ and therefore suffers from the same gradient starvation on the prompt axis (Theorem~\ref{thm:starvation}).

\subsection{Additional Ablation Studies}
\label{app:additional_ablations}

\subsubsection{Full LLM Results (All Methods)}

Table~\ref{tab:additional_summary_full} expands the main-body Table~\ref{tab:additional_summary} to show all five training methods (Base, SFT, DPO, GRPO, Multi-DPOP) with AlpacaEval-2 LC and MMLU-Pro. The pattern is consistent: SFT and DPO improve standard accuracy and Inverse-IFEval modestly but leave abductive accuracy near base. GRPO provides the strongest forward baseline but still falls far short of Multi-DPOP on both abductive benchmarks. General capability (MMLU-Pro) is preserved across all methods except SFT, which slightly degrades due to training on curated completions only.

\begin{table}[htbp]
\centering
\caption{Full comparison of all baselines and methods across five LLM families. \textbf{Acc.}: HaluEval; \textbf{A-Acc.}: A-HaluEval; \textbf{Inv-IF}: Inverse-IFEval accuracy; \textbf{LC}: AlpacaEval-2 LC; \textbf{MMLU}: MMLU-Pro.}
\vspace{2pt}
\label{tab:additional_summary_full}
\resizebox{\linewidth}{!}{%
\begin{tabular}{llccccc}
\toprule
\textbf{Model} & \textbf{Method} & \textbf{Acc.} & \textbf{A-Acc.} & \textbf{Inv-IF} & \textbf{LC} & \textbf{MMLU} \\
\midrule
\multirow{5}{*}{Tulu-3.1-8B}
  & Base       & $90.0$ & $54.7$ & $32.8$ & $34.5$  & $20.2$ \\
  & SFT        & $93.2$ & $55.8$ & $35.1$  & $33.1$  & $19.6$ \\
  & DPO        & $96.8$ & $57.1$ & $38.4$  & $36.7$  & $20.0$ \\
  & GRPO       & $97.5$ & $58.3$ & $40.2$  & $36.3$  & $20.0$ \\
  & Multi-DPOP & $\mathbf{99.5}$ & $\mathbf{85.0}$ & $\mathbf{65.3}$ & $36.9$ & $20.3$ \\
\midrule
\multirow{5}{*}{Gemma-3-4B}
  & Base       & $87.5$ & $52.3$ & $38.5$ & $41.2$  & $43.6$ \\
  & SFT        & $91.4$ & $53.5$ & $40.2$  & $39.8$  & $42.9$ \\
  & DPO        & $94.6$ & $55.2$ & $42.1$  & $42.5$  & $43.4$ \\
  & GRPO       & $95.8$ & $56.1$ & $43.7$  & $42.8$  & $43.4$ \\
  & Multi-DPOP & $\mathbf{98.5}$ & $\mathbf{82.5}$ & $\mathbf{69.8}$ & $43.1$ & $43.7$ \\
\midrule
\multirow{5}{*}{GLM-4-9B}
  & Base       & $88.3$ & $53.8$ & $42.1$ & $62.3$  & $27.7$ \\
  & SFT        & $92.1$ & $54.9$ & $44.8$  & $60.7$  & $27.0$ \\
  & DPO        & $95.4$ & $56.3$ & $47.3$  & $63.5$  & $27.5$ \\
  & GRPO       & $96.2$ & $57.4$ & $48.5$  & $63.8$  & $27.5$ \\
  & Multi-DPOP & $\mathbf{99.0}$ & $\mathbf{84.0}$ & $\mathbf{74.2}$ & $64.2$ & $27.8$ \\
\midrule
\multirow{5}{*}{Qwen3-4B}
  & Base       & $89.1$ & $55.2$ & $44.7$ & $56.2$ & $50.6$ \\
  & SFT        & $92.8$ & $56.4$ & $46.3$  & $54.5$ & $49.8$ \\
  & DPO        & $96.1$ & $58.0$ & $49.8$  & $57.2$ & $50.4$ \\
  & GRPO       & $97.2$ & $59.3$ & $51.2$  & $57.5$ & $50.4$ \\
  & Multi-DPOP & $\mathbf{99.5}$ & $\mathbf{86.5}$ & $\mathbf{76.1}$ & $58.4$ & $50.7$ \\
\midrule
\multirow{5}{*}{Qwen3-30B}
  & Base       & $91.2$ & $57.6$ & $49.2$ & $67.4$ & $61.5$ \\
  & SFT        & $94.5$ & $58.9$ & $51.4$  & $65.3$ & $60.7$ \\
  & DPO        & $97.3$ & $60.2$ & $55.1$  & $68.5$ & $61.3$ \\
  & GRPO       & $98.1$ & $61.4$ & $56.8$  & $68.9$ & $61.3$ \\
  & Multi-DPOP & $\mathbf{100.0}$ & $\mathbf{91.5}$ & $\mathbf{83.7}$ & $69.7$ & $\mathbf{61.6}$ \\
\bottomrule
\end{tabular}
}
\end{table}

\subsubsection{Inverse-IFEval Detailed Results}

Table~\ref{tab:inverse_ifeval} presents the full Inverse-IFEval results alongside IFBench (general instruction following). Two key findings emerge: (i)~A-DPO and Multi-DPOP substantially improve Inverse-IFEval accuracy ($62$--$84\%$), surpassing closed-source models including Claude-4-Opus ($67.2\%$) and GPT-5-high ($73.7\%$); and (ii)~DPO significantly degrades IFBench ($-8$ to $-12\%$), indicating overfitting to constraint patterns, while A-DPO and Multi-DPOP preserve IFBench within $\pm 0.5\%$.

\begin{table}[htbp]
\centering
\caption{Results on \textsc{Inverse-IFEval} and \textsc{IFBench}. Inv-IF: Inverse-IFEval accuracy (higher = better at counter-conventional constraints). IFB: IFBench accuracy (general instruction following; higher = better). DPO significantly degrades IFBench; GRPO slightly degrades it; A-DPO and Multi-DPOP preserve IFBench while improving Inverse-IFEval.}
\vspace{2pt}
\label{tab:inverse_ifeval}
\resizebox{\linewidth}{!}{
\begin{tabular}{lcccccccccc}
\toprule
& \multicolumn{2}{c}{\textbf{Tulu-3.1-8B}} & \multicolumn{2}{c}{\textbf{Gemma-3-4B}} & \multicolumn{2}{c}{\textbf{GLM-4-9B}} & \multicolumn{2}{c}{\textbf{Qwen3-4B}} & \multicolumn{2}{c}{\textbf{Qwen3-30B}} \\
\cmidrule(lr){2-3} \cmidrule(lr){4-5} \cmidrule(lr){6-7} \cmidrule(lr){8-9} \cmidrule(lr){10-11}
\textbf{Method} & Inv-IF & IFB & Inv-IF & IFB & Inv-IF & IFB & Inv-IF & IFB & Inv-IF & IFB \\
\midrule
\multicolumn{11}{c}{\textit{Closed-Source Models (Zero-Shot)}} \\
\midrule
GPT-5-high     & \multicolumn{2}{c}{$73.7$} & \multicolumn{8}{c}{} \\
Gemini-2.5-Pro & \multicolumn{2}{c}{$70.6$} & \multicolumn{8}{c}{} \\
Claude-4-Opus  & \multicolumn{2}{c}{$67.2$} & \multicolumn{8}{c}{} \\
\midrule
\multicolumn{11}{c}{\textit{Open-Source Models}} \\
\midrule
Base Model & $32.8$ & $82.4$ & $38.5$ & $76.3$ & $42.1$ & $69.0$ & $44.7$ & $80.5$ & $49.2$ & $83.9$ \\
\midrule
SFT        & $35.1$ & $78.6$ & $40.2$ & $72.5$ & $44.8$ & $65.3$ & $46.3$ & $76.8$ & $51.4$ & $80.1$ \\
DPO        & $38.4$ & $\mathit{71.2}$ & $42.1$ & $\mathit{64.8}$ & $47.3$ & $\mathit{57.4}$ & $49.8$ & $\mathit{69.3}$ & $55.1$ & $\mathit{73.6}$ \\
GRPO       & $40.2$ & $79.8$ & $43.7$ & $73.6$ & $48.5$ & $66.1$ & $51.2$ & $77.9$ & $56.8$ & $81.3$ \\
\midrule
A-DPO      & $62.7$ & $82.1$ & $67.4$ & $76.0$ & $71.8$ & $68.7$ & $73.6$ & $80.2$ & $80.5$ & $83.5$ \\
Multi-DPOP & $\mathbf{65.3}$ & $\mathbf{82.6}$ & $\mathbf{69.8}$ & $\mathbf{76.5}$ & $\mathbf{74.2}$ & $\mathbf{69.2}$ & $\mathbf{76.1}$ & $\mathbf{80.7}$ & $\mathbf{83.7}$ & $\mathbf{84.1}$ \\
\bottomrule
\end{tabular}
}
\end{table}

\subsubsection{Lambda Ablation}

\begin{figure}[htbp]
\centering
\includegraphics[width=0.75\linewidth]{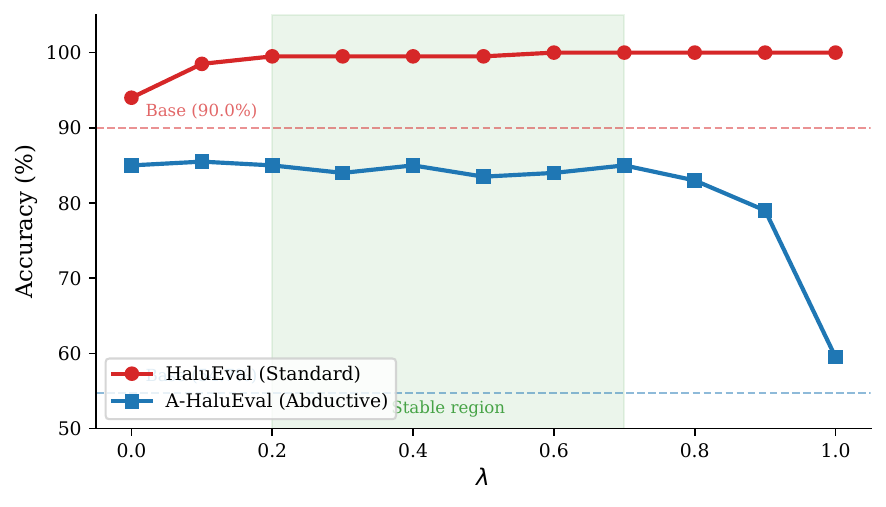}
\caption{Effect of $\lambda$ on \textsc{HaluEval} (Tulu-3.1-8B). Dashed lines: base model. The shaded region ($\lambda \in [0.2, 0.7]$) shows stable performance on both axes.}
\label{fig:lambda_ablation}
\end{figure}

Figure~\ref{fig:lambda_ablation} shows the effect of the multitask weight $\lambda$ in $\mathcal{L}_{\text{Multi-DPO}}$ (\Cref{eq:multi}). Performance is stable across $\lambda \in [0.2, 0.7]$: both standard HaluEval accuracy ($>99\%$) and abductive A-HaluEval accuracy ($>83\%$) remain high. Beyond $\lambda > 0.7$, the forward objective dominates and abductive accuracy drops sharply to $59.5\%$ at $\lambda = 1.0$ (pure DPO). We use $\lambda = 0.5$ throughout as a safe default.

\subsubsection{Data Efficiency}

Table~\ref{tab:data_ablation} shows A-DPO performance on VLMBias when trained with 25\%, 50\%, 75\%, and 100\% of the training data (Qwen3-VL-8B). Even with only $\sim\!212$ pairs (25\%), A-DPO achieves $25.3\%$ — already $2\times$ above the best forward baseline (GRPO at $11.8\%$ with full data). Performance scales smoothly from $25.3\%$ to $43.5\%$, indicating the method learns genuine visual grounding rather than memorizing specific image pairs.

\begin{table}[htbp]
\centering
\caption{Data efficiency ablation on \textsc{VLMBias} (Qwen3-VL-8B, A-DPO). Even 25\% of training data outperforms all forward baselines.}
\vspace{2pt}
\label{tab:data_ablation}
\resizebox{0.5\linewidth}{!}{%
\begin{tabular}{lcc}
\toprule
\textbf{Data fraction} & \textbf{HumorDB} & \textbf{VLMBias} \\
\midrule
25\%  & $72.4$ & $25.3$ \\
50\%  & $78.1$ & $33.8$ \\
75\%  & $82.3$ & $39.4$ \\
100\% & $85.0$ & $43.5$ \\
\bottomrule
\end{tabular}
}
\end{table}

\subsubsection{Multi-DPOP Penalty ($\lambda_{\text{DPOP}}$)}
Table~\ref{tab:dpop_ablation} ablates the DPOP penalty weight $\lambda_{\text{DPOP}}$ at fixed $\lambda = 0.5$ (\Cref{eq:multi}). Standard accuracy remains at $100\%$ across all values, while abductive accuracy varies modestly ($87$--$91\%$). Lower $\lambda_{\text{DPOP}}$ slightly favors abductive accuracy, but the effect is small ($\leq 4\%$), showing that the DPOP penalty is not a critical hyperparameter for the abductive learning framework.
\begin{table*}[ht]
\centering
\caption{Ablation studies for the penalty $\lambda_{\text{DPOP}}$ of the original preference learning objective ($\lambda = 0.5$, \eqref{eq:multi}), utilizing the \textsc{A-HaluEval} dataset generated via gpt-4o.}
\vspace{1pt}
\label{tab:dpop_ablation}
{
\begin{tabular}{ccc}
\toprule
{$\lambda_{\text{DPOP}}$}  &{\bf Accuracy} (\textsc{HaluEval}, $\%$) & {\bf Accuracy} (\textsc{A-HaluEval}, $\%$)\\
\midrule
$1.0$ & $100.0$ & $87.0$ \\
$0.9$ & $100.0$ & $87.0$ \\
$0.8$ & $100.0$ & $87.5$ \\
$0.7$ & $100.0$ & $87.5$  \\
$0.6$ & $100.0$ & $87.5$ \\
$0.5$ & $100.0$ & $88.5$  \\
$0.4$ & $100.0$ & $90.0$ \\
$0.3$ & $100.0$ & $89.5$ \\
$0.2$ & $100.0$ & $90.0$ \\
$0.1$ & $100.0$ & $91.0$  \\
$0.0$ & $100.0$ & $91.0$  \\
\bottomrule
\end{tabular}
}
\end{table*}

\subsection{Notation Summary}
\label{app:notation}

Table~\ref{tab:notation} summarizes the distributions used throughout the paper: population quantities derived from the marginal $p$, and training-time quantities set by the data pipeline, derived from the within-tuple sampling $p_{\mathcal{S}}$.

\begin{table}[htbp]
\centering
\caption{Notation for distributions. Top: population quantities (properties of the world). Bottom: training-time quantities (properties of the data pipeline).}
\vspace{2pt}
\label{tab:notation}
\resizebox{\linewidth}{!}{%
\begin{tabular}{lp{9.2cm}}
\toprule
\textbf{Symbol} & \textbf{Meaning} \\
\midrule
$p(x)$ & Population prompt marginal; carries the rarity premise $p(x_s) \ll p(x_g)$ (Theorems~\ref{thm:starvation},~\ref{thm:amplification}). \\
$q_\pi(y) := \sum_{x'} \pi(y \mid x')\, p(x')$ & Response marginal induced by policy $\pi$; instances $q_\theta$, $q_{\mathrm{ref}}$; subscript dropped when the policy is clear. \\
$\widetilde{\pi}(x \mid y) = \pi(y \mid x)\, p(x) / q(y)$ & Population abductive conditional, induced from the forward policy via Bayes' rule (Eq.~\ref{eq:abductive_pi}); written $\pi(x \mid y)$ where unambiguous. \\
$\mathcal{D}_g, \mathcal{D}_s$ & General/specific subsets of the forward training data ($N_g \gg N_s$ examples, inheriting the population imbalance). \\
\midrule
$\mathcal{D}$ & Curated abductive preference dataset of triples $(x_w, x_l, y)$ (Theorem~\ref{prop:general}). \\
$p_{\mathcal{S}}(x)$ & Within-tuple candidate sampling distribution at training time; uniform ($1/K$) by construction of the curation pipeline (Corollary~\ref{cor:infonce}). \\
$q_{\mathcal{S}}(y) := \sum_j p_{\mathcal{S}}(x_j)\, \pi_\theta(y \mid x_j)$ & Response marginal induced by $p_{\mathcal{S}}$. \\
$\pi_{\mathcal{S}}(x \mid y) \propto p_{\mathcal{S}}(x)\, \pi_\theta(y \mid x)$ & Uniform-prior abductive conditional; the target of the swapped InfoNCE (Corollary~\ref{cor:infonce}) and the quantity measured by abductive accuracy (Appendix~\ref{app:metrics}). \\
\bottomrule
\end{tabular}
}
\end{table}

\subsection{Proofs of Main Theoretical Results}
\label{app:proofs}

Throughout, let $f_\theta: \mathbb{R}^d \to \mathbb{R}$ be a twice-differentiable function with parameter-input mixed Hessian $\mathbf{H}_\theta(x) := \frac{\partial^2 f_\theta}{\partial \theta \, \partial x^T}\big|_x \in \mathbb{R}^{p \times d}$.
The dataset contains $N_g$ general samples with $x_i = v_c$ and $N_s$ specific samples with $x_i = v_c + v_d$, where $\|v_d\|$ is small.
We write $\epsilon := \|v_d\|$ and track all remainder terms in $O(\epsilon^2)$.

\subsubsection{Proof of Theorem~\ref{thm:starvation} (Detail Suppression in Pre-Training)}

\begin{proof}
\textbf{Step 1: Per-position gradient decomposition.}
The autoregressive log-likelihood decomposes as $\log \pi_\theta(y \mid x) = \sum_{t=1}^{T} \log \pi_\theta(y_t \mid x, y_{<t})$.
The gradient of the pre-training loss $\mathcal{L} = -\sum_i \log \pi_\theta(y_i \mid x_i)$ is:
\begin{equation}
\label{eq:app_mle_grad}
    \nabla_\theta \mathcal{L} = -\sum_{i=1}^{N_g + N_s} \sum_{t=1}^{T} \nabla_\theta \log \pi_\theta(y_{i,t} \mid x_i, y_{i,<t}).
\end{equation}
Each per-position term $\nabla_\theta \log \pi_\theta(y_t \mid x, y_{<t})$ is a function of $x$ through the model's hidden states.
We decompose the sum by sample type.

\textbf{Step 2: General samples ($x_i = v_c$).}
All general samples share $x_i = v_c$, so their per-position gradients are identical:
\begin{equation}
    \mathbf{G}_c := -\sum_{i \in \mathcal{D}_g} \sum_{t=1}^{T} \nabla_\theta \log \pi_\theta(y_{g,t} \mid v_c, y_{g,<t}) = -N_g \cdot \nabla_\theta \log \pi_\theta(y_g \mid v_c).
\end{equation}
This is $O(N_g)$ and depends only on $v_c$, carrying no information about $v_d$.

\textbf{Step 3: Specific samples ($x_i = v_c + v_d$).}
For each position $t$, apply Taylor expansion of $\nabla_\theta \log \pi_\theta(y_t \mid x, y_{<t})$ around $x = v_c$:
\begin{equation}
    \nabla_\theta \log \pi_\theta(y_t \mid v_c + v_d, y_{<t}) = \nabla_\theta \log \pi_\theta(y_t \mid v_c, y_{<t}) + \mathbf{J}_t(v_c, y_{<t}) \, v_d + O(\|v_d\|^2), \label{eq:app_grad_taylor}
\end{equation}
where $\mathbf{J}_t(x, y_{<t}) := \frac{\partial^2 \log \pi_\theta(y_t \mid x, y_{<t})}{\partial \theta\, \partial x^T}$ is the per-position mixed Hessian.
Summing over all positions:
\begin{align}
    \nabla_\theta \log \pi_\theta(y_s \mid v_c + v_d) &= \nabla_\theta \log \pi_\theta(y_s \mid v_c) + \bar{\mathbf{H}}_\theta(v_c, y_s) \, v_d + O(\|v_d\|^2), \label{eq:app_Gs}
\end{align}
where $\bar{\mathbf{H}}_\theta(x, y) := \sum_{t=1}^{T} \mathbf{J}_t(x, y_{<t})$ is the aggregate mixed Hessian.

\textbf{Step 4: Identifying common and detail components.}
The specific-sample contribution is:
\begin{align}
    \mathbf{G}_s &:= -\sum_{i \in \mathcal{D}_s} \nabla_\theta \log \pi_\theta(y_i \mid x_i) \nonumber \\
    &= -N_s \nabla_\theta \log \pi_\theta(y_s \mid v_c) \quad &&\text{(zeroth order: common direction)} \nonumber \\
    &\quad -\; N_s \, \bar{\mathbf{H}}_\theta(v_c, y_s) \, v_d \quad &&\text{(first order: \textbf{detail direction})} \nonumber \\
    &\quad +\; O(\|v_d\|^2). &&\label{eq:app_Gs_expanded}
\end{align}
The total gradient $\nabla_\theta \mathcal{L} = \mathbf{G}_c + \mathbf{G}_s$ has detail component $\bar{\mathbf{H}}_\theta(v_c, y_s) \, v_d$ appearing only in $\mathbf{G}_s$ ($N_s$ terms), while the common component $\nabla_\theta \log \pi_\theta(\cdot \mid v_c)$ appears in both $\mathbf{G}_c$ and $\mathbf{G}_s$ ($N_g + N_s$ terms).

\textbf{Step 5: Gradient ratio bound.}
Let $\mathbf{d} := \bar{\mathbf{H}}_\theta(v_c, y_s)\, v_d / \|\bar{\mathbf{H}}_\theta(v_c, y_s)\, v_d\|$ denote the unit detail direction.
Under $N_g \gg N_s$:
\begin{equation}
\label{eq:app_ratio}
    \frac{|\langle \nabla_\theta \mathcal{L},\, \mathbf{d} \rangle|}{\|\nabla_\theta \mathcal{L}\|} = O\!\left(\frac{N_s}{N_g + N_s}\right) \xrightarrow{N_g/N_s \to \infty} 0.
\end{equation}

\textbf{Step 6: Scalar special case.}
When $T = 1$ and the model reduces to a scalar logit $f_\theta(x)$ with per-sample loss $\ell(f_\theta(x), y)$, the aggregate Hessian reduces to $\bar{\mathbf{H}}_\theta = \ell''_g \cdot \mathbf{H}_\theta(v_c)$ where $\mathbf{H}_\theta(x) = \frac{\partial^2 f_\theta}{\partial \theta\, \partial x^T}$, recovering the standard scalar-model result.
In the linear MSE case $f_\theta(x) = W^T x$ with weight decay $\lambda$, the stationary point satisfies $w_c^* \approx 1$ and $w_d^* = N_s \delta / (N_s + \lambda)$.
\end{proof}

\subsubsection{Proof of Theorem~\ref{thm:forward_balanced} (Detail-Blindness of Forward Preference Learning)}
\label{app:forward_balanced}

\begin{proof}
\textbf{Step 1: Per-sample score gradient.}
Each forward sample has $x_i = v_c + v_d$ (all samples are specific under balanced data). The score difference is $s_i = \psi(x_i, y_{w,i}) - \psi(x_i, y_{l,i})$, with gradient:
\begin{equation}
    \nabla_\theta s_i = \nabla_\theta \log \pi_\theta(y_{w,i} \mid v_c + v_d) - \nabla_\theta \log \pi_\theta(y_{l,i} \mid v_c + v_d).
\end{equation}

\textbf{Step 2: Taylor expansion of each term.}
Expanding around $v_c$:
\begin{align}
    \nabla_\theta \log \pi_\theta(y \mid v_c + v_d) &= \nabla_\theta \log \pi_\theta(y \mid v_c) + \bar{\mathbf{H}}_\theta(v_c, y) \cdot v_d + O(\|v_d\|^2),
\end{align}
where $\bar{\mathbf{H}}_\theta(v_c, y)$ is the aggregate mixed Hessian from Theorem~\ref{thm:starvation}.

\textbf{Step 3: Subtraction.}
\begin{align}
    \nabla_\theta s_i &= \underbrace{\Big[\nabla_\theta \log \pi_\theta(y_{w,i} \mid v_c) - \nabla_\theta \log \pi_\theta(y_{l,i} \mid v_c)\Big]}_{\text{Term A: } O(1),\; v_d\text{-blind}} + \underbrace{\Big[\bar{\mathbf{H}}_\theta(v_c, y_{w,i}) - \bar{\mathbf{H}}_\theta(v_c, y_{l,i})\Big] v_d}_{\text{Term B: } O(\|v_d\|),\; v_d\text{-sensitive}} + O(\|v_d\|^2). \label{eq:app_forward_balanced_decomp}
\end{align}
Term~A is the gradient of the response preference evaluated at the \emph{general} prompt $v_c$; it is $O(1)$ and independent of $v_d$.
Term~B captures how the response preference changes as the prompt moves from $v_c$ to $v_c + v_d$; it is $O(\|v_d\|)$.

\textbf{Step 4: Full loss gradient.}
The loss gradient $\nabla_\theta \mathcal{L}_{\mathrm{fwd}} = -\sum_i F'(s_i) \cdot \nabla_\theta s_i$ is:
\begin{equation}
    \nabla_\theta \mathcal{L}_{\mathrm{fwd}} = -\sum_{i=1}^{N} F'(s_i) \cdot [\text{Term A}_i] \;-\; \sum_{i=1}^{N} F'(s_i) \cdot [\text{Term B}_i] + O(\|v_d\|^2).
\end{equation}
Since $F'(s_i)$ is a scalar that multiplies both terms of the \emph{same} sample identically, it cannot change the ratio of Term~A to Term~B within any sample. The per-sample detail fraction $\|\text{Term B}_i\| / (\|\text{Term A}_i\| + \|\text{Term B}_i\|) = O(\|v_d\|)$ is invariant to the choice of $F$ (DPO, IPO, SLiC, or any other pairwise loss).

\textbf{Step 5: Contrast with backward training.}
Under backward data $(x_w, x_l, y)$ with $x_w - x_l = v_d$, the score difference is $s_i = \psi(x_{w,i}, y_i) - \psi(x_{l,i}, y_i)$, giving:
\begin{equation}
    \nabla_\theta s_i = \nabla_\theta \log \pi_\theta(y_i \mid v_c + v_d) - \nabla_\theta \log \pi_\theta(y_i \mid v_c) = \bar{\mathbf{H}}_\theta(v_c, y_i) \cdot v_d + O(\|v_d\|^2).
\end{equation}
The zeroth-order terms $\nabla_\theta \log \pi_\theta(y_i \mid v_c)$ cancel exactly because the same response $y_i$ appears with both prompts.
The entire gradient is in the $v_d$ direction; the per-sample detail fraction is $1$.
\end{proof}

\subsubsection{Proof of Theorem~\ref{thm:amplification} (Inverse Amplification)}
\label{app:proof_amplification}

\begin{proof}
By Bayes' rule, $\pi'(y_s \mid x_s) = \pi'(x_s \mid y_s) \cdot q'(y_s) / p(x_s)$.
The perturbed response marginal satisfies
\begin{equation}
    q'(y_s) = q(y_s) + \delta \cdot p(x_s),
\end{equation}
since only the $x_s$ term in $q(y_s) = \sum_{x'} \pi(x' \mid y_s)\, p(x')$ is affected by the perturbation $\pi'(x_s \mid y_s) = \pi(x_s \mid y_s) + \delta$.
Substituting:
\begin{align}
    \pi'(y_s \mid x_s) &= \frac{(\pi(x_s \mid y_s) + \delta)(q(y_s) + \delta\, p(x_s))}{p(x_s)} \nonumber \\
    &= \frac{\pi(x_s \mid y_s) \cdot q(y_s)}{p(x_s)} + \delta \cdot \frac{q(y_s)}{p(x_s)} + \delta \cdot \pi(x_s \mid y_s) + \frac{\delta^2 \cdot p(x_s)}{p(x_s)}.
\end{align}
The first term is $\pi(y_s \mid x_s)$.
Taking the difference:
\begin{equation}
    \Delta_{\pi(y_s \mid x_s)} = \delta \cdot \frac{q(y_s)}{p(x_s)} + \delta \cdot \pi(x_s \mid y_s) + \delta^2.
\end{equation}
The dominant term is $\delta \cdot q(y_s)/p(x_s)$.
The remaining terms are $O(\delta)$ since $\pi(x_s \mid y_s) \leq 1$ and $\delta$ is small, giving the stated result.
\end{proof}

\subsubsection{Proof of Theorem~\ref{prop:general} (Role-Switch Equivalence)}
\label{app:prop}

\begin{proof}
\textbf{Step 1: Abductive policy via Bayes' theorem.}
By \eqref{eq:abductive_pi}, the abductive policies for $\pi_\theta$ and $\pi_{\mathrm{ref}}$ are:
\begin{align}
    \widetilde{\pi}_\theta(x \mid y) &= \frac{\pi_\theta(y \mid x)\, p(x)}{q_\theta(y)}, &
    \widetilde{\pi}_{\mathrm{ref}}(x \mid y) &= \frac{\pi_{\mathrm{ref}}(y \mid x)\, p(x)}{q_{\mathrm{ref}}(y)}, \label{eq:app_abd_policies}
\end{align}
where $q_\theta(y) = \sum_{x'} \pi_\theta(y \mid x')\, p(x')$ and similarly for $q_{\mathrm{ref}}$.

\textbf{Step 2: Log-ratio decomposition.}
The abductive comparison score is:
\begin{align}
    \widetilde{\psi}(x, y) &:= \log \frac{\widetilde{\pi}_\theta(x \mid y)}{\widetilde{\pi}_{\mathrm{ref}}(x \mid y)} \nonumber \\
    &= \log \frac{\pi_\theta(y \mid x)\, p(x) / q_\theta(y)}{\pi_{\mathrm{ref}}(y \mid x)\, p(x) / q_{\mathrm{ref}}(y)} \nonumber \\
    &= \log \frac{\pi_\theta(y \mid x)}{\pi_{\mathrm{ref}}(y \mid x)} + \log \frac{q_{\mathrm{ref}}(y)}{q_\theta(y)} \nonumber \\
    &= \psi(x, y) + C(y), \label{eq:app_psi_decomp}
\end{align}
where $\psi(x, y) := \log \pi_\theta(y \mid x) / \pi_{\mathrm{ref}}(y \mid x)$ is the forward score, and $C(y) := \log q_{\mathrm{ref}}(y)/q_\theta(y)$ depends on $y$ alone.
Crucially, $C(y)$ does not depend on $x$ because condition~(a) ensures both models share the same prompt marginal $p(x)$.

\textbf{Step 3: Translation invariance.}
The abductive loss with outer function $F$ is:
\begin{align}
    \widetilde{\mathcal{L}}(\theta) &= -\mathbb{E}_{(x_w, x_l, y)\sim\mathcal{D}}\Big[F\big(\widetilde{\psi}(x_w, y),\, \widetilde{\psi}(x_l, y)\big)\Big] \nonumber \\
    &= -\mathbb{E}_{(x_w, x_l, y)\sim\mathcal{D}}\Big[F\big(\psi(x_w, y) + C(y),\, \psi(x_l, y) + C(y)\big)\Big]. \label{eq:app_abd_loss_expanded}
\end{align}
By condition~(b), $F$ is translation-invariant: $F(a + c, b + c) = F(a, b)$ for all $c \in \mathbb{R}$.
Applying this with $c = C(y)$:
\begin{equation}
    \widetilde{\mathcal{L}}(\theta) = -\mathbb{E}_{(x_w, x_l, y)\sim\mathcal{D}}\Big[F\big(\psi(x_w, y),\, \psi(x_l, y)\big)\Big]. \label{eq:app_role_switch}
\end{equation}
This is the role-switched loss: pairwise learning of $\pi(x \mid y)$ using the \emph{forward} score $\psi$ on swapped data.

\textbf{Step 4: Corollary for DPO.}
For DPO, $F(a, b) = \log \sigma(\beta(a - b))$. Translation invariance holds since $F(a+c, b+c) = \log\sigma(\beta(a+c-b-c)) = \log\sigma(\beta(a-b)) = F(a,b)$.

\textbf{Step 5: Extension to IPO, GPO, and SLiC-HF.}
Similarly, the $F$ function for IPO is  $-(a - b - \frac{1}{2\beta})^2$, for GPO is $\rho^{-1}(a - b)$, where $\rho$ is a general monotone link, and for SLiC-HF is $\min(0, a - b - \gamma)$, being translation invariant.
\end{proof}

\subsubsection{Proof of Theorem~\ref{thm:recovery} (Gradient Recovery)}
\label{sec:gradient_recovery}

\begin{proof}
\textbf{Step 1: Backward pairwise score gradient.}
The backward score difference is:
\begin{equation}
    s_i := \psi(x_{w,i}, y_i) - \psi(x_{l,i}, y_i).
\end{equation}
Its gradient with respect to $\theta$ is:
\begin{equation}
    \nabla_\theta s_i = g(x_w, y_i) \cdot \nabla_\theta f_\theta(x_w) - g(x_l, y_i) \cdot \nabla_\theta f_\theta(x_l). \label{eq:app_bwd_si}
\end{equation}

\textbf{Step 2: Taylor expansion of the coupling function.}
Since $x_w = x_l + v_d$ and $\|v_d\| = \epsilon$ is small:
\begin{equation}
    g(x_w, y_i) = g(x_l, y_i) + \nabla_x g(x_l, y_i)^T v_d + O(\epsilon^2) =: g_l + O(\epsilon). \label{eq:app_g_taylor}
\end{equation}

\textbf{Step 3: Taylor expansion of the parameter gradient.}
By \eqref{eq:app_grad_taylor}:
\begin{equation}
    \nabla_\theta f_\theta(x_w) = \nabla_\theta f_\theta(x_l) + \mathbf{H}_\theta(x_l)\, v_d + O(\epsilon^2). \label{eq:app_bwd_grad_taylor}
\end{equation}

\textbf{Step 4: Substitution and cancellation.}
Substituting \eqref{eq:app_g_taylor} and \eqref{eq:app_bwd_grad_taylor} into \eqref{eq:app_bwd_si}:
\begin{align}
    \nabla_\theta s_i &= \big[g_l + O(\epsilon)\big] \cdot \big[\nabla_\theta f_\theta(x_l) + \mathbf{H}_\theta(x_l)\, v_d + O(\epsilon^2)\big] - g_l \cdot \nabla_\theta f_\theta(x_l) \nonumber \\
    &= g_l \cdot \nabla_\theta f_\theta(x_l) + g_l \cdot \mathbf{H}_\theta(x_l)\, v_d - g_l \cdot \nabla_\theta f_\theta(x_l) + O(\epsilon^2) \nonumber \\
    &= g_l \cdot \mathbf{H}_\theta(x_l)\, v_d + O(\epsilon^2). \label{eq:app_bwd_cancellation}
\end{align}
The zeroth-order term $g_l \cdot \nabla_\theta f_\theta(x_l)$ \textbf{cancels exactly}.
What remains is the first-order term $g_l \cdot \mathbf{H}_\theta(x_l)\, v_d$, which is proportional to the detail feature $v_d$.

\textbf{Step 5: Full backward gradient.}
The total backward gradient is:
\begin{align}
    \nabla_\theta \mathcal{L}_{\mathrm{bwd}} &= -\sum_{i=1}^{N} F'(s_i) \cdot \nabla_\theta s_i \nonumber \\
    &= -\sum_{i=1}^{N} F'(s_i) \cdot g(x_{l,i}, y_i) \cdot \mathbf{H}_\theta(x_{l,i})\, v_d + O(\epsilon^2). \label{eq:app_bwd_full}
\end{align}
Every sample ($N$ total) contributes a gradient aligned with $\mathbf{H}_\theta(x_l)\, v_d$.
There is no common-feature component at first order; it was eliminated by the contrastive subtraction in Step~4.

\textbf{Step 6: Comparison with forward.}
\begin{itemize}
    \item \textbf{Forward (Theorem~\ref{thm:forward_balanced}):} Per-sample detail fraction is $O(\|v_d\|) \ll 1$, even when all prompts are specific.
    \item \textbf{Backward (this theorem):} Detail component involves $N$ terms; common component cancelled. Ratio $= \infty$ (detail dominates entirely).
\end{itemize}

\textbf{Step 7: Linear special case.}
For $f_\theta(x) = W^T x$: $g(x, y) = (y - W^T x)/\sigma^2$, $\nabla_\theta f = x$, $\mathbf{H}_\theta = I$.
The cancellation in Step~4 becomes exact (no $O(\epsilon^2)$ remainder).
Projecting onto $v_d$: $\partial s_i / \partial w_d = (y_i - w_c - w_d)/\sigma^2 \approx \delta/\sigma^2$.
Projecting onto $v_c$: $\partial s_i / \partial w_c = -w_d/\sigma^2 \approx 0$.
\end{proof}

% \subsubsection{Proof of Corollary~\ref{cor:data_balance} (Data Balance Requirement)}
% \begin{proof}
% From Theorem~\ref{thm:forward_pairwise}, the detail gradient magnitude is $O(N_s \cdot |\overline{F' \Delta g}|_s \cdot \|\mathbf{H}_\theta v_d\|)$ and the common gradient magnitude is $O((N_g + N_s) \cdot |\overline{F' \Delta g}| \cdot \|\nabla_\theta f(v_c)\|)$.
% For comparable magnitudes: $N_s \gtrsim N_g \cdot \|\nabla_\theta f(v_c)\| / \|\mathbf{H}_\theta v_d\|$.
% In the linear case ($\nabla_\theta f = x$, $\mathbf{H}_\theta = I$), this reduces to $N_s \gtrsim N_g \cdot \|v_c\| / \|v_d\|$. Since $\|v_d\|$ is small, the requirement is even more stringent than near-equal representation.
% \end{proof}

\subsubsection{Proof of Corollary~\ref{cor:dpop_bound} (Bounded Approximation for A-DPOP)}
\label{app:dpop_bound}

\begin{proof}
The true abductive loss uses the abductive score $\widetilde{\psi}(x, y) = \psi(x, y) + C(y)$ (cf.\ proof of Theorem~\ref{prop:general}):
\begin{equation}
    \widetilde{\mathcal{L}}_{\mathrm{DPOP}} = -\mathbb{E}\!\left[h\big(\psi(x_w, y) - \psi(x_l, y)\big) + r\big(\psi(x_w, y) + C(y)\big)\right].
\end{equation}
The $h$ term is translation-invariant, so $C(y)$ cancels exactly as in Theorem~\ref{prop:general}.
The practical role-switched loss uses $r(\psi(x_w, y))$ without $C(y)$:
\begin{equation}
    \mathcal{L}_{\mathrm{A\text{-}DPOP}} = -\mathbb{E}\!\left[h\big(\psi(x_w, y) - \psi(x_l, y)\big) + r\big(\psi(x_w, y)\big)\right].
\end{equation}
The difference is:
\begin{align}
    \big|\widetilde{\mathcal{L}}_{\mathrm{DPOP}} - \mathcal{L}_{\mathrm{A\text{-}DPOP}}\big|
    &= \big|\mathbb{E}\!\left[r(\psi(x_w, y) + C(y)) - r(\psi(x_w, y))\right]\big| \nonumber \\
    &\leq \lambda_{\mathrm{DPOP}} \cdot \mathbb{E}\!\left[\big|\max(0, -\psi(x_w, y) - C(y)) - \max(0, -\psi(x_w, y))\big|\right] \nonumber \\
    &\leq \lambda_{\mathrm{DPOP}} \cdot \mathbb{E}\!\left[|C(y)|\right],
\end{align}
where the last step uses the 1-Lipschitz continuity of $\max(0, \cdot)$.
At initialization $\theta = \theta_{\mathrm{ref}}$, $q_\theta = q_{\mathrm{ref}}$ and $C(y) = 0$ for all $y$, so the bound is exactly zero.
During fine-tuning, $\mathbb{E}[|C(y)|]$ grows as $\theta$ diverges from $\theta_{\mathrm{ref}}$, but remains small when fine-tuning is performed on small datasets with limited parameter updates.
\end{proof}

\subsubsection{Proof of Corollary~\ref{cor:infonce} (Extension to Contrastive Losses)}
\label{app:infonce}

\begin{proof}
Let the training joint be $p_{\mathcal{S}}(x, y) := p_{\mathcal{S}}(x)\,\pi_\theta(y \mid x)$, where $p_{\mathcal{S}}$ is uniform over the candidate pool $\{x_1, \ldots, x_K\}$, i.e.\ $p_{\mathcal{S}}(x_j) = 1/K$ for each $j$. Let $x^+ = x_1$ denote the positive. The swapped InfoNCE loss is
\begin{equation}
    \mathcal{L}_{\mathrm{swap}} \;=\; -\log \frac{\pi_\theta(y \mid x^+)}{\sum_{j=1}^{K} \pi_\theta(y \mid x_j)}.
\end{equation}
The response marginal induced by the training joint is
\begin{equation}
    q_{\mathcal{S}}(y) \;=\; \sum_{j=1}^K p_{\mathcal{S}}(x_j)\,\pi_\theta(y \mid x_j) \;=\; \frac{1}{K}\sum_{j=1}^K \pi_\theta(y \mid x_j),
\end{equation}
and Bayes' rule applied to $p_{\mathcal{S}}$ gives the abductive conditional
\begin{equation}
    \pi_{\mathcal{S}}(x^+ \mid y) \;=\; \frac{p_{\mathcal{S}}(x^+)\,\pi_\theta(y \mid x^+)}{q_{\mathcal{S}}(y)} \;=\; \frac{(1/K)\,\pi_\theta(y \mid x^+)}{(1/K)\sum_{j=1}^K \pi_\theta(y \mid x_j)} \;=\; \frac{\pi_\theta(y \mid x^+)}{\sum_{j=1}^K \pi_\theta(y \mid x_j)}.
\end{equation}
Substituting back,
\begin{equation}
    \mathcal{L}_{\mathrm{swap}} \;=\; -\log \pi_{\mathcal{S}}(x^+ \mid y),
\end{equation}
which is the cross-entropy of classifying the positive prompt $x^+$ under the abductive conditional $\pi_{\mathcal{S}}(x \mid y)$. Within-tuple uniformity is therefore sufficient for the role-switch identity to hold exactly, independent of any property of the population marginal $p$.
\end{proof}

\paragraph{Relation to the population conditional.}
The target $\pi_{\mathcal{S}}(x \mid y)$ is the posterior under the uniform within-tuple prior. Relative to the population conditional $\widetilde{\pi}(x \mid y) \propto \pi_\theta(y \mid x)\, p(x)$, the per-tuple log-odds differ by exactly the population prior log-ratio: with $p_{\mathcal{S}}$ uniform,
\begin{equation*}
    \log \frac{\pi_{\mathcal{S}}(x_w \mid y)}{\pi_{\mathcal{S}}(x_l \mid y)} \;=\; \log \frac{\widetilde{\pi}(x_w \mid y)}{\widetilde{\pi}(x_l \mid y)} \;-\; \log \frac{p(x_w)}{p(x_l)}.
\end{equation*}
Under the rarity premise of Theorem~\ref{thm:starvation} this prior ratio is large for minimally different prompts, so the two conditionals are distinct objectives. Training and evaluation both use the uniform-prior target: it removes the memorized prompt prior that abductive training overrides (Corollary~\ref{cor:infonce}), and the abductive-accuracy metric (Appendix~\ref{app:metrics}) measures the same comparison.

\subsection{Evaluation Metric Definitions}
\label{app:metrics}

\paragraph{Accuracy (HaluEval).} The model's capability of choosing the right answer given the original prompt via pairwise log-probability comparison:
$\pi(y_{\text{right}} \mid x) > \pi(y_{\text{hallucinated}} \mid x)$.

\paragraph{Abductive Accuracy (A-HaluEval).} The model's capability of preferring the correct prompt given a shared response:
$\pi(y \mid x_w) > \pi(y \mid x_l)$.
This is the $K{=}2$ comparison under the uniform-prior abductive conditional of Corollary~\ref{cor:infonce}: $\pi_{\mathcal{S}}(x_w \mid y) > \pi_{\mathcal{S}}(x_l \mid y) \iff \pi(y \mid x_w) > \pi(y \mid x_l)$.
The uniform prior keeps the memorized bias $p(x)$ out of the measurement: the metric asks which prompt better explains the response, the quantity whose gradient is analyzed in Theorem~\ref{thm:recovery}.

\paragraph{Inverse-IFEval accuracy.} Generation-based: we sample responses and evaluate whether the model correctly follows counter-conventional constraints using deterministic rule-based parsing (same as IFEval~\citep{zhou2023instructionfollowingevaluationlargelanguage} — no LLM judge required). We also report \textsc{IFBench} accuracy to measure whether training degrades general instruction following.

\paragraph{HumorDB / VLMBias.} Generation-based: the model generates a free-form answer (``Is this funny?'', ``How many legs?'') and accuracy is computed against ground truth.

\paragraph{General capability.} MMLU-Pro (5-shot), GPQA Diamond (0-shot), AlpacaEval-2 LC, SimpleQA, and TruthfulQA for LLMs; MMMU-Pro and HallusionBench for VLMs.

\subsection{Training Details per Dataset}
\label{app:training_details}
This section provides the detailed dataset-specific training procedures summarized in Section~4.

\subsubsection{\textsc{HaluEval} QA}
We construct abductive triples $(x_w, x_l, y)$ by minimally editing background knowledge while keeping the question fixed, so that the shared response $y$ is supported only under $x_w$ (Table~\ref{tab:abductive_halueval}). We evaluate (i) in-domain on the \textsc{HaluEval} test split, (ii) factual QA robustness on \textsc{SimpleQA} and \textsc{TruthfulQA}, and (iii) general knowledge capability on \textsc{MMLU-Pro}, \textsc{GPQA}, and \textsc{AlpacaEval-2}.

\subsubsection{\textsc{Inverse-IFEval}}
\textsc{Inverse-IFEval}~\citep{zhang2025inverseifevalllmsunlearn} targets a common failure mode of instruction-tuned LLMs: they can become \emph{stubbornly} tied to familiar instruction-following patterns learned during training (e.g., IFEval-style constraints~\citep{zhou2023instructionfollowingevaluationlargelanguage}) and fail to generalize when the instruction is phrased in an unusual or counterintuitive way. We evaluate (i) in-domain on the \textsc{Inverse-IFEval} test split and (ii) out-of-domain on \textsc{IFBench}~\cite{pyatkin2025generalizing} to examine whether training degrades general instruction-following capability.

\subsubsection{\textsc{HumorDB}}
\textsc{HumorDB}~\citep{jain2024ai} is a curated benchmark for visual humor with minimally contrastive image pairs (photos, cartoons, sketches, and AI-generated images) where subtle edits flip whether an image is perceived as funny. We construct abductive preference data by attaching the identical text prompt \textit{``Is this image funny?''} to each humorous/non-humorous image pair and using the shared response $y{=}$\textit{``Yes''}. The humorous image (with the prompt) is treated as $x_w$ and the non-humorous image as $x_l$, so the model must attend to the visual differences that flip the label. We merge the training and validation splits for training ($991$ pairs) and use the test split for evaluation ($300$ pairs).

\subsubsection{\textsc{VLMBias}}
\textsc{VLMBias}~\citep{vo2025visionlanguagemodelsbiased} targets \emph{objective} counterfactual visual perception, where models often default to memorized priors (e.g., ``dogs have 4 legs'') instead of grounding in the image. We construct abductive pairs by keeping the text prompt fixed (e.g., ``How many legs does this dog have?'') and setting $y$ to the counterfactual-correct answer (e.g., ``5''), with the counterfactual image as $x_w$ and the original image as $x_l$. We use $847$ counterfactual image pairs for training and $200$ for evaluation.

\subsection{Training Hyperparameters for Reproducibility}
\label{app:hyperparameters}
All experiments are conducted on 8$\times$H100 GPUs using DeepSpeed ZeRO Stage 2 via \texttt{accelerate launch}.

\subsubsection{Text Models}
Table~\ref{tab:hyper_text} summarizes the training hyperparameters for text-based experiments. \textsc{HaluEval} (H) uses larger batches with higher per-device batch size, while \textsc{Inverse-IFEval} (IF) uses gradient accumulation to compensate for longer prompt sequences. Multi-DPOP uses a higher learning rate for the combined loss.

\begin{table}[htbp]
\centering
\caption{Training hyperparameters for text models. H\,=\,HaluEval; IF\,=\,Inverse-IFEval.}
\vspace{2pt}
\label{tab:hyper_text}
\resizebox{\linewidth}{!}{%
\begin{tabular}{lcccc}
\toprule
\textbf{Hyperparameter} & \textbf{H (A-DPO)} & \textbf{H (Multi-DPOP)} & \textbf{IF (A-DPO)} & \textbf{IF (Multi-DPOP)} \\
\midrule
\texttt{max\_length} & 8192 & 8192 & 8192 & 8192 \\
\texttt{max\_prompt\_length} & 4096 & 4096 & 2048 & 2048 \\
\texttt{batch size} (per device) & 16 & 16 & 1 & 1 \\
\texttt{gradient\_accumulation} & 2 & 2 & 16 & 8 \\
\texttt{num\_epochs} & 5 & 5 & 10 & 5 \\
\texttt{learning\_rate} & $5\!\times\!10^{-7}$ & $5\!\times\!10^{-7}$ & $5\!\times\!10^{-7}$ & $5\!\times\!10^{-6}$ \\
\texttt{lr\_scheduler} & const+warmup & const+warmup & const+warmup & const+warmup \\
\texttt{warmup\_ratio} & 0.1 & 0.1 & 0.1 & 0.1 \\
\texttt{max\_grad\_norm} & 1.0 & 1.0 & 1.0 & 1.0 \\
$\beta$ & 0.05 & 0.05 & 0.05 & 0.05 \\
$\lambda$ (Multi) & -- & 0.5 & -- & 0.5 \\
$\lambda_{\text{DPOP}}$ & -- & 0.1 & -- & 0.1 \\
\texttt{optimizer} & \multicolumn{4}{c}{Adam ($\beta_1{=}0.9,\;\beta_2{=}0.999,\;\varepsilon{=}10^{-8}$, weight decay $= 0.0$)} \\
\texttt{LoRA} & \multicolumn{4}{c}{None (full fine-tuning)} \\
\bottomrule
\end{tabular}
}
\end{table}

\subsubsection{Vision-Language Models}
Table~\ref{tab:hyper_vlm} summarizes the hyperparameters for VLM experiments. All VLM models use full fine-tuning. \textsc{HumorDB} uses shorter sequences than \textsc{VLMBias} since humor captions are brief, while the image resolution cap (\texttt{max\_pixels}) is identical for both to keep visual token counts tractable across two images per ADPO sample.

\begin{table}[htbp]
\centering
\caption{Training hyperparameters for VLMs (A-DPO).}
\vspace{2pt}
\label{tab:hyper_vlm}
\resizebox{0.85\linewidth}{!}{%
\begin{tabular}{lcc}
\toprule
\textbf{Hyperparameter} & \textbf{VLMBias} & \textbf{HumorDB} \\
\midrule
\texttt{max\_length} & 8192 & 2048 \\
\texttt{max\_prompt\_length} & 2048 & 1024 \\
\texttt{max\_pixels} & $262{,}144\;(512{\times}512)$ & $262{,}144\;(512{\times}512)$ \\
\texttt{batch size} (per device) & 1 & 1 \\
\texttt{gradient\_accumulation} & 16 & 16 \\
\texttt{num\_epochs} & 3 & 3 \\
\texttt{learning\_rate} & $5\!\times\!10^{-7}$ & $5\!\times\!10^{-7}$ \\
\texttt{lr\_scheduler} & cosine & cosine \\
\texttt{warmup\_ratio} & 0.1 & 0.1 \\
\texttt{max\_grad\_norm} & 1.0 & 1.0 \\
$\beta$ & 0.1 & 0.1 \\
\texttt{optimizer} & \multicolumn{2}{c}{AdamW ($\beta_1{=}0.9,\;\beta_2{=}0.999$, weight decay $= 0.01$)} \\
\bottomrule
\end{tabular}
}
\end{table}

\end{document}